\documentclass[journal]{IEEEtran}
\usepackage{tikz}
\usepackage{hyperref}
\usepackage[switch, pagewise]{lineno}
\usepackage{xcolor}
\usepackage{units}
\usepackage{graphicx}
\usepackage{amssymb}
\usepackage{cite}
\usepackage{graphicx}
\usepackage{epstopdf}
\usepackage{subfig}
\usepackage{array}
\usepackage{amsmath}
\usepackage{url}
\usepackage{pdflscape}
\usepackage{rotating,booktabs}
\usepackage{multirow}
 \usepackage{gensymb}
\usepackage{caption}
\usepackage[noabbrev]{cleveref}
\usepackage{rotating} % for vertical text in table cells
\usepackage{ colortbl }% for color cell
 \usepackage{esvect}
\usepackage{mathtools}

\newenvironment{blue}{\color{black}}{}

\newenvironment{orange}{\color{black}}{}

\newcommand{\verText}[1]{\begin{turn}{-60}{#1}\end{turn}}
\newcommand{\beginsupplement}{%
        \setcounter{table}{0}
        \renewcommand{\thetable}{S\arabic{table}}%
        \setcounter{figure}{0}
        \renewcommand{\thefigure}{S\arabic{figure}}%
     }
\DeclarePairedDelimiter\floor{\lfloor}{\rfloor}

\iffalse
\newcommand\copyrighttext{%
  \centering Copyright (c) 2016 IEEE. Personal use of this material is permitted. However, permission to use this material for any other purposes must be obtained from the IEEE by sending a request to pubs-permissions@ieee.org}
\newcommand\copyrightnotice{%
\begin{tikzpicture}[remember picture,overlay]
\node[anchor=south,yshift=10pt] at (current page.south) {{\parbox{\dimexpr\textwidth-\fboxsep-\fboxrule\relax}{\copyrighttext}}};
\end{tikzpicture}%
}
\else
\newcommand\copyrighttext{%
  \centering {\footnotesize An accepted version of N. Kajbakhsh, J. Y. Shin, S. Gurudu, R. T. Hurst, C. B. Kendall, M. B. Gotway, and J. Liang. ``Convolutional neural networks for medical image analysis: Full training or fine tuning?'' {\em IEEE Transactions on Medical Imaging.} 35(5):1299-1312, 2016}}
\newcommand\copyrightnotice{%
\begin{tikzpicture}[remember picture,overlay]
\node[anchor=south,yshift=10pt] at (current page.south) {{\parbox{\dimexpr\textwidth-\fboxsep-\fboxrule\relax}{\copyrighttext}}};
\end{tikzpicture}%
}
\fi

\begin{document}

%%%%%%%%%%%%%%%%%%%%%%%%Responses
%%%%%%%%%%%%%%%%%%%%%%%%Responses
\onecolumn
\setcounter{figure}{0}
\renewcommand{\thefigure}{\Alph{figure}}
%\include{ResponseLetter_Rev2}
%%%%%%%%%%%%%%%%%%%%%%%%Responses
%%%%%%%%%%%%%%%%%%%%%%%%Responses

        \setcounter{figure}{0}
        \renewcommand{\thefigure}{\arabic{figure}}

\twocolumn
\setlength{\parindent}{1.0em}
\setlength{\parskip}{0ex plus 0.2ex minus 0.1ex}
\title{Convolutional Neural Networks for Medical Image Analysis: Full Training or Fine Tuning?}
\author{Nima Tajbakhsh$^{*}$,~\IEEEmembership{Member,~IEEE,} Jae Y. Shin$^{*}$, Suryakanth R. Gurudu,  R. Todd Hurst, Christopher B. Kendall, Michael B. Gotway, and Jianming Liang,~\IEEEmembership{Senior Member,~IEEE}% <-this % stops a space
\thanks{N.~Tajbakhsh, J.~Y.~Shin,  and J.~Liang are with the Department
of Biomedical Informatics, Arizona State University, 13212 East Shea Boulevard, Scottsdale, AZ 85259, USA (e-mail: \{Nima.Tajbakhsh, Sejong, Jianming.Liang\}@asu.edu). Nima Tajbakhsh and Jae Y. Shin have contributed equally.}% <-this % stops a space
\thanks{S.~R.~Gurudu (Division of Gastroenterology and Hepatology); R.~T.~Hurst, C.~Kendall (Division of Cardiovascular Diseases); and M.~B.~Gotway (Department of Radiology) are with Mayo Clinic, 13400 E. Shea Blvd., Scottsdale, AZ 85259, USA (e-mail: \{Hurst.R, Kendall.Christopher, Gurudu.Suryakanth, Gotway.Michael\}@mayo.edu).}% <-this % stops a space
}

\maketitle
{\blue
\begin{abstract}
Training a deep convolutional neural network (CNN) from scratch is difficult because it requires a large amount of labeled training data and a great deal of expertise to ensure proper convergence. A promising alternative is to fine-tune a CNN that has been pre-trained using, for instance, a large set of labeled natural images. However, the substantial differences between natural and medical images may advise against such knowledge transfer. In this paper, we seek to answer the following central question in the context of medical image analysis: \emph{Can the use of pre-trained deep CNNs with sufficient fine-tuning eliminate the need for training a deep CNN from scratch?} To address this question, we considered 4 distinct medical imaging applications in 3 specialties (radiology, cardiology, and gastroenterology) involving classification, detection, and segmentation from 3 different imaging modalities, and investigated how the performance of deep CNNs trained from scratch compared with the pre-trained CNNs fine-tuned in a layer-wise manner. Our experiments consistently demonstrated that (1) the use of a pre-trained CNN with adequate fine-tuning outperformed or, in the worst case, performed as well as a CNN trained from scratch; (2) fine-tuned CNNs were more robust to the size of training sets than CNNs trained from scratch; (3) neither shallow tuning nor deep tuning was the optimal choice for a particular application; and (4) our layer-wise fine-tuning scheme could offer a practical way to reach the best performance for the application at hand based on the amount of available data. 
\end{abstract}
}


\IEEEpeerreviewmaketitle
\copyrightnotice

\section{Introduction}
\label{intro}

Convolutional neural networks (CNNs) have been used in the field of computer vision for decades \cite{fukushima80,le98,le15}. However, their true value had not been discovered until the ImageNet competition in 2012, a success that brought about a revolution through the efficient use of graphics processing units (GPUs), rectified linear units, new dropout regularization, and effective data augmentation \cite{le15}. Acknowledged as one of the top 10 breakthroughs of 2013 \cite{breakthrough}, CNNs have once again become a popular learning machine, now not only within the computer vision community but across various applications ranging from natural language processing to hyperspectral image processing and to medical image analysis. The main power of a CNN lies in its deep architecture \cite{szegedy14,simonyan14,zeiler14,Eigen13}, which allows for extracting a set of discriminating features at multiple levels of abstraction.

However, training a deep CNN from scratch (or full training) is not without complications \cite{erhan09}. First, CNNs require a large amount of labeled training data---a requirement that may be difficult to meet in the medical domain where expert annotation is expensive and the diseases (e.g., lesions) are scarce in the datasets. Second, training a deep CNN requires extensive computational and memory resources, without which the training process would be extremely time-consuming. Third, training a deep CNN is often complicated by overfitting and convergence issues, whose resolution frequently requires repetitive adjustments in the architecture or learning parameters of the network to ensure that all layers are learning with comparable speed. Therefore, deep learning from scratch can be tedious and time-consuming, demanding a great deal of diligence, patience, and expertise. 

A promising alternative to training a CNN from scratch is to fine-tune a CNN that has been trained using a large labeled dataset from a different application. The pre-trained models have been applied successfully to various computer vision tasks as a feature generator or as a baseline for transfer learning \cite{razavian14,azizpour14,penatti}. Herein, we address the following central question in the context of medical image analysis:    \emph{Can the use of pre-trained deep CNNs with sufficient fine-tuning eliminate the need for training a deep CNN from scratch?} This is an important question because training deep CNNs from scratch may not be practical, given the limited labeled data in medical imaging. To answer this central question, we conducted an extensive set of experiments for 4 medical imaging applications: 1) polyp detection in colonoscopy videos, 2) image quality assessment in colonoscopy videos, 3) pulmonary embolism detection in computed tomography (CT) images, and 4) intima-media boundary segmentation in ultrasonographic images. We have chosen these applications to represent the most common clinically used imaging modality systems (i.e., CT, ultrasonography, and optical endoscopy) and the most common medical image analysis tasks (i.e., lesion detection, image segmentation, and image classification). For each application, we compared the performance of the pre-trained CNNs through fine-tuning with that of the CNNs trained from scratch entirely based on medical imaging data. We also compared the performance of the CNN-based systems with their corresponding handcrafted counterparts.  

%{\red Our extensive set of experiments in Section~\ref{apps} collectively demonstrated that the depth or level of fine-tuning was essential to successful knowledge transfer. We have observed incremental performance improvement with deeper fine-tuning of the pre-trained CNNs. With sufficient labeled training data, pre-trained CNNs with adequate fine-tuning outperform or, in the worst case, perform as well as the CNNs trained from scratch. With reduced training data, fine-tuned CNNs tended to outperform to an even greater degree the CNNs trained from scratch. These results suggest that fine-tuned CNNs should be the preferred choice, even with ample amount of labeled data available. Furthermore, for each application studied in this paper, we have observed that both the CNN trained from scratch and the sufficiently fine-tuned CNN surpass the corresponding handcrafted approach by a relatively large margin, further confirming the potential of CNNs in medical imaging applications.}

\section{Related Works}

Applications of CNNs in medical image analysis can be traced to the 1990s, when they were used for computer-aided detection of microcalcifications in digital mammography \cite{zhang94,chan95} and computer-aided detection of lung nodules in CT datasets \cite{lo95}. With revival of CNNs owing to the development of powerful GPU computing, the medical imaging literature has witnessed a new generation of computer-aided detection systems that show superior performance. Examples include automatic polyp detection in colonoscopy videos \cite{tajbakhsh15b}\cite{tajbakhsh15a}, computer-aided detection of pulmonary embolism (PE) in CT datasets \cite{tajbakhsh15c}, automatic detection of mitotic cells in histopathology images \cite{cirecsan13},  computer-aided detection of lymph nodes in CT images \cite{roth14}, and computer-aided anatomy detection in CT volumes \cite{zheng15}. Applications of CNNs in medical image analysis are not limited to only computer-aided detection systems, however. CNNs have recently been used for carotid intima-media thickness measurement in ultrasound images \cite{shin16}, pancreas segmentation in CT images \cite{roth15a},  brain tumor segmentation in magnetic resonance imaging (MRI) scans \cite{havaei15}, multimodality isointense infant brain image segmentation \cite{zhang15}, neuronal membrane segmentation in electron microscopy images \cite{ciresan12}, and knee cartilage segmentation in MRI scans\cite{prasoon13}.

One important aspect of CNNs is the ``transferability'' of knowledge embedded in the pre-trained CNNs. Recent research conducted by Azizpour et al. \cite{azizpour14} suggests that the success of knowledge transfer depends on the distance, or dissimilarity, between the database on which a CNN is trained and the database to which the knowledge is to be transferred. Although the distance between natural image and medical imaging databases is considerable, recent studies show the potential for knowledge transfer to the medical imaging domain.  

{\orange The recent research on transfer learning in medical imaging can be categorized into two groups. The first group \cite{bar15, Ginneken15, Arevalo15} consists of works  wherein a pre-trained CNN is used as a feature generator. Specifically, the pre-trained CNN is applied to an input image and then the CNN outputs (features)  are extracted from a certain layer of the network. The extracted features are then used to train a new pattern classifier. For instance, in \cite{bar15}, pre-trained CNNs  were used as a feature generator for chest pathology identification. A similar study \cite{Ginneken15} by Ginneken et al. showed that although the use of pre-trained CNNs could not outperform a dedicated nodule detection system, the integration of CNN-based features with the handcrafted features enabled improved performance. 

The second group \cite{schlegl14, chen15,  Gustavo15,shin15,Gao15,margeta15} consists of works  wherein a pre-trained CNN is adapted to the application at hand. For instance, in \cite{Gustavo15}, the fully connected layers of a pre-trained CNN were replaced with a new logistic layer, and then the labeled data were used to train only the appended layer while keeping the rest of the network the same. This treatment yielded promising results for classification of unregistered multiview mammogram. Chen et al. \cite{chen15} suggested the use of a fine-tuned pre-trained CNN for localizing standard planes in ultrasound images. {\blue In \cite{Gao15}, the authors  fine-tuned all layers of a pre-trained CNN for automatic classification of interstitial lung diseases. They also suggested an attenuation rescale scheme to convert 1-channel CT slices to RGB-like images needed for tuning the pre-trained model. Shin et al. \cite{shin15} used fine-tuned pre-trained CNNs to automatically map medical images to document-level topics, document-level sub-topics, and sentence-level topics. In \cite{margeta15}, fine-tuned pre-trained CNNs were used to automatically retrieve missing or noisy cardiac acquisition plane information from magnetic resonance imaging and predict the five most common cardiac views.}  Different from the previous approaches, Schlegl et al. \cite{schlegl14} considered the fine-tuning of an unsupervised network. They explored unsupervised pre-training of CNNs to inject information from sites or image classes for which no annotations were available, and showed that such across site pre-training improved classification accuracy compared to random initialization of the model parameters.}

\section{Contributions}

In this paper, we systematically study knowledge transfer to medical imaging applications, making the following contributions:

\begin{itemize}

{\orange
\item We demonstrated how fine-tuning a pre-trained CNN in a layer-wise manner leads to incremental performance improvement. This approach distinguishes our work from \cite{bar15, Ginneken15, Arevalo15}, which downloaded the features from the fully connected layers of a pre-trained CNN and then trained a separate pattern classifier. Our approach also differs from \cite{schlegl14, chen15,  Gustavo15} wherein the entire pre-trained CNN underwent fine-tuning. 

\item We analyzed how the availability of training samples influences the choice between pre-trained CNNs and CNNs trained from scratch. To our knowledge, this issue has not yet been systematically addressed in the medical imaging literature. 

\item We compared the performance of pre-trained CNNs, not only against handcrafted approaches but also against CNNs trained from scratch using medical imaging data. This analysis is in contrast to \cite{bar15, Ginneken15}, who provided only limited performance comparisons between pre-trained CNNs and handcrafted approaches.

\item We presented consistent results with conclusive outcomes for 4 distinct medical imaging applications involving classification, detection, and segmentation in 3 different medical imaging modalities, which add substantially to the state of the art where conclusions are based solely on 1 medical imaging application.
}

\end{itemize}

\section{Convolutional Neural Networks (CNNs)}
\label{CNN}

CNNs are so-named because of the convolutional layers in their architectures. Convolutional layers are responsible for detecting certain local features in all locations of their input images. To detect local structures, each node in a convolutional layer is connected to only a small subset of spatially connected neurons in the input image channels. To enable the search for the same local feature throughout the input channels, the connection weights are shared between the nodes in the convolutional layers. Each set of shared weights is called a \textit{kernel}, or a \textit{convolution kernel}. Thus, a convolutional layer with $n$ kernels learns to detect $n$ local features whose strength across the input images is visible in the resulting $n$ feature maps. To reduce computational complexity and achieve a hierarchical set of image features, each sequence of convolution layers is followed by a \textit{pooling layer}, a workflow reminiscent of simple and complex cells in the primary visual cortex \cite{hubel59}. The max pooling layer reduces the size of feature maps by selecting the maximum feature response in overlapping or non-overlapping local neighborhoods, discarding the exact location of such maximum responses. As a result, max pooling can further improve translation invariance. CNNs typically consist of several pairs of convolutional and pooling layers, followed by a number of consecutive  fully connected layers, and finally a \textit {softmax layer}, or \textit { regression layer}, to generate the desired outputs. In more modern CNN architectures, computational efficiency is achieved by replacing the pooling layer with a convolution layer with a stride larger than 1.

Similar to multilayer perceptrons, CNNs are trained with the back-propagation algorithm by minimizing the following cost function with respect to the unknown weights $W$:

{\orange
\begin{equation}
\mathcal{L} = -\frac{1}{|X|}\sum_{i}^{|X|}{\ln(p(y^i|X^i))}
\label{costFunc}
\end{equation}
  
\noindent where $|X|$ denotes the number of training images,}  $X^i$ denotes the $i^{th}$ training image with the corresponding label $y^i$, and $p(y^i|X^i)$ denotes the probability by which $X^i$ is correctly classified. Stochastic gradient descent is commonly used for minimizing this cost function, where the cost over the entire training set is approximated with the cost over mini-batches of data. If $W_l^t$ denotes the weights in $l^{th}$ convolutional layer at iteration $t$, and $\mathcal{\hat{L}}$ denotes the cost over a mini-batch of size $N$, then the updated weights in the next iteration are computed as follows:

$$\gamma^t=\gamma^{\floor*{\frac{tN}{|X|}}}$$
$$V_l^{t+1} = \mu V_l^t - \gamma^t\alpha_l \frac{\partial \mathcal{\hat{L}}}{\partial W_l}$$
\begin{equation}
W_l^{t+1} = W_l^t + V_l^{t+1}
\label{eq:weightUpdate}
\end{equation}

\noindent where $\alpha_l$ is the learning rate of the $l^{th}$ layer, $\mu$ is the momentum that indicates the contribution of the previous weight update in the current iteration, and $\gamma$ is the scheduling rate that decreases learning rate $\alpha$ at the end of each epoch. 

\begin{table*}
\caption{The AlexNet architecture used in our experiments. Of note, $C$ is the number of classes, which is 3 for intima-media interface segmentation and is 2 for colonoscopy frame classification, polyp detection, and pulmonary embolism detection.}
\begin{center}
\begin{tabular}{*{7}{|c|c|c|c|c|c|c}}

\cline{1-7}
layer&type&input&kernel&stride&pad&output\\
\hline
%\parbox[c]{3mm}{\multirow{8}{*}{\rotatebox[origin=c]{90}{ \# voting groups}}}&   $K$=2&  0.863 & $0.905$ &  $0.921$ &  0.928 & 0.928\\[.1cm]
data& input  & 3x227x227 & N/A & N/A &N/A & 3x227x227 \\[.1cm]
conv1& convolution  & 3x227x227 & 11x11 & 4 & 0 & 96x55x55\\[.1cm]
pool1& max pooling  & 96x55x55 & 3x3 & 2 & 0 & 96x27x27\\[.1cm]
conv2& convolution  & 96x27x27 & 5x5 & 1 & 2 & 256x27x27\\[.1cm]
pool2& max pooling  & 256x27x27 & 3x3 & 2 & 0 & 256x13x13\\[.1cm]
conv3& convolution  & 256x13x13 & 3x3 & 1 & 1 & 384x13x13\\[.1cm]
conv4& convolution  & 384x13x13 & 3x3 & 1 & 1 & 384x13x13\\[.1cm]
conv5& convolution  & 384x13x13 & 3x3 & 1 & 1 & 256x13x13\\[.1cm]
pool5& max pooling  & 256x13x13 & 3x3 & 2 & 0 & 256x6x6\\[.1cm]
fc6& fully connected  & 256x6x6 & 6x6 & 1 & 0 & 4096x1\\[.1cm]
fc7& fully connected  & 4096x1 & 1x1 & 1 & 0 & 4096x1\\[.1cm]
fc8& fully connected  & 4096x1 & 1x1 & 1 & 0 & Cx1\\[.1cm]
\hline
\end{tabular}
\end{center}
\label{table:netArch}
\end{table*}

\section{Fine-tuning}
\label{tuning}
The iterative weight update in Eq.~\ref{eq:weightUpdate} begins with a set of randomly initialized weights. Specifically, before the commencement of the training phase, weights in each convolutional layer of a CNN are initialized by values randomly sampled from a normal distribution with a zero mean and small standard deviation. However, considering the large number of weights in a CNN and the limited availability of labeled data, the iterative weight update, starting with a random weight initialization, may lead to an undesirable local minimum for the cost function. Alternatively, the weights of the convolutional layers can be initialized with the weights of a pre-trained CNN with the same architecture. The pre-trained net is generated with a massive set of labeled data from a different application. Training a CNN from a set of pre-trained weights is called \textit{fine-tuning} and has been used successfully in several applications \cite{razavian14,azizpour14,penatti}.

{\orange Fine-tuning begins with copying (transferring) the weights from a pre-trained network to the network we wish to train. The exception is the last fully connected layer whose number of nodes depends on the number of classes in the dataset. A common practice is to replace the last fully connected layer of the pre-trained CNN with a new fully connected layer that  has as many neurons as the number of classes in the new target application.  In our study, we deal with 2-class and 3-class classification tasks; therefore,  the new fully connected layer has 2 or 3 neurons depending on the application under study. After the weights of the last fully connected layer are initialized, the new network can be fine-tuned in a layer-wise manner, starting with tuning only the last layer, then tuning all layers in a CNN.}

 Consider a CNN with $L$ layers where the last 3 layers are fully connected layers. Also let $\alpha_l$ denote the learning rate of the $l^{th}$ layer in the network. We can fine-tune only the last (new) layer of the network by setting $\alpha_l=0$ for $l\neq L$. This level of fine-tuning corresponds to training a linear classifier with the features generated in layer $L-1$. Likewise, the last 2 layers of the network can be fine-tuned by setting $\alpha_l=0$ for $l\neq L, L-1$. This level of fine-tuning corresponds to training an artificial neural network with 1 hidden layer, which can be viewed as training a nonlinear classifier using the features generated in layer $L-2$. Similarly, fine-tuning layers $L$, $L-1$, and $L-2$ are essentially equivalent to training an artificial neural network with 2 hidden layers. Including the previous convolution layers in the update process further adapts the pre-trained CNN to the application at hand but may require more labeled training data to avoid overfitting.

{\orange  In general, the early layers of a CNN learn low level image features, which are applicable to most vision tasks, but the late layers learn high-level features, which are specific to the application at hand. Therefore, fine-tuning the last few layers is usually sufficient for transfer learning. However, if the distance between the source and target applications is significant, one may need to fine-tune the early layers as well. Therefore, an effective fine-tuning technique is to start from the last layer and then incrementally include more layers in the update process until the desired performance is reached.} We refer to tuning the last few convolutional layers as ``shallow tuning" and we consider tuning all the convolutional layers as ``deep tuning". 
{\orange  We would like to note that the suggested fine-tuning scheme differs from \cite{razavian14,penatti} wherein the network remains the same and serves as a feature generator, and also differs from \cite{azizpour14} wherein the entire network undergoes fine-tuning at once. }

\begin{table*}
\caption{Learning parameters used for training and fine-tuning of AlexNet in our experiments. $\mu$ is the momentum, $\alpha$ is the learning rate of the weights in each convolutional layer, and $\gamma$ determines how $\alpha$ decreases over epochs. The learning rate for the bias term is always set twice as large as the learning rate of the corresponding weights. Of note, ``fine-tuned AlexNet:layer1-layer2" indicates that all layers between and including these 2 layers undergo fine-tuning.}
\begin{center}
\begin{tabular}{*{11}{|c|c|c|c|c|c|c|c|c|c|c}}

\cline{1-11}
\multirow{ 2}{*}{CNNs} &\multicolumn{10}{|c|}{Parameters} \\
\cline{2-11}
&$\mu$&$\alpha_{conv1}$&$\alpha_{conv2}$&$\alpha_{conv3}$&$\alpha_{conv4}$&$\alpha_{conv5}$&$\alpha_{fc6}$&$\alpha_{fc7}$&$\alpha_{fc8}$&$\gamma$\\
\hline
 
%\multirow{ 2}{*}{\multicolumn{10}{c}{PE detection}} \\

Fine-tuned AlexNet:conv1-fc8 &0.9&0.001&0.001&0.001&0.001&0.001&0.001&0.001&0.01&0.95\\
Fine-tuned AlexNet:conv2-fc8 &0.9&0&0.001&0.001&0.001&0.001&0.001&0.001&0.01&0.95\\
Fine-tuned AlexNet:conv3-fc8 &0.9&0& 0&0.001&0.001&0.001&0.001&0.001&0.01&0.95\\
Fine-tuned AlexNet:conv4-fc8 &0.9&0&0&0&0.001&0.001&0.001&0.001&0.01&0.95\\
Fine-tuned AlexNet:conv5-fc8 &0.9&0&0&0&0&0.001&0.001&0.001&0.01&0.95\\
Fine-tuned AlexNet:fc6-fc8 &0.9&0&0&0&0&0&0.001&0.001&0.01&0.95\\
Fine-tuned AlexNet:fc7-fc8 &0.9&0&0&0&0&0&0&0.001&0.01&0.95\\
Fine-tuned AlexNet:only fc8 &0.9&0&0&0&0&0&0&0&0.01&0.95\\
AlexNet scratch &0.9&0.001&0.001&0.001&0.001&0.001&0.001&0.001&0.001&0.95\\

\hline
\end{tabular}
\end{center}
\label{table:netParams}
\end{table*}

\section{Applications and Results}
\label{apps}

{\orange In our study, we considered 4 different medical imaging applications from 3 imaging modality systems. We study the performance of polyp detection and PE detection using a free-response operating characteristic (FROC) analysis, analyze the performance of frame classification by means of an ROC analysis, and evaluate the performance of boundary segmentation through a boxplot analysis.  To perform statistical comparisons, we have computed the  error bars  corresponding to 95\% confidence intervals for both ROC and FROC curves according to the method suggested in \cite{Edwards02}. The error bars enable us to compare each pair of performance curves at multiple operating points from a statistical perspective. Specifically, if the error bars of a pair of curves do not overlap at a fixed false positive rate, then the two curves are statistically different (p$<$.05) at the given operating point. An appealing feature of this statistical analysis is that we can compare the performance curves at a clinically acceptable operating point rather than comparing the curves as a whole. While we have discussed the statistical comparisons throughout the paper, we have further summarized them in a number of tables in supplementary material, which can be found in the supplementary files/multimedia tab.}

 We used the Caffe library \cite{jia14} for both training and fine-tuning CNNs. For consistency and ease of comparison, we used the AlexNet architecture for the 4 applications under study.  {\orange  Training and fine-tuning of each AlexNet took approximately  2-3  hours  depending on the size of the training set. {\blue To ensure the  proper convergence of each CNN, we monitored the area under the receiver operating characteristic curve. Specifically, for each experiment, we divided the training set into a smaller training set with 80\% of the training data and a validation set with the remaining 20\% of the training data and then computed area under the curve on the validation set. The training process was terminated when the highest accuracy on the validation set was observed.} All  training  was  performed  using an  NVIDIA  GeForce  GTX 980TI (6GB  on-board  memory). The fully trained CNNs were initialized with random weights sampled from  Gaussian distributions. We also experimented with other initialization techniques such as those suggested in \cite{glorot10} and  \cite{he15}, but we observed no significant performance gain after convergence, even though we noticed varying speed of convergence using these initialization techniques.}

{\orange  For both full training and fine-tuning scenarios, we used a stratified training set of image patches where the positive and negative classes were equally present. For this purpose, we randomly down-sampled the majority (negative) class,  while keeping the minority class (positive) unchanged.} For the fine-tuning scenario, we used the pre-trained AlexNet model provided in the Caffe library. {\orange  The pre-trained AlexNet consists of approximately 5 million parameters in the convolution layers and about 55 million parameters in its fully connected layers, and is trained using  1.2 million images labeled with 1000 semantic classes. The model used in our study is the snapshot taken after 360,000 training iterations.} As shown in Table~\ref{table:netArch}, AlexNet begins with 2 pairs of convolutional and pooling layers, mapping the 227x227 input images to 13x13 feature maps. This architecture then proceeds with a sequence of 3 convolutional layers that efficiently implement a convolutional layer with 9x9 kernels, yet with a larger degree of nonlinearity. The sequence of convolutional layers is then followed by a pooling layer and 3 fully connected layers. The first fully connected layer can be viewed as a convolution layer with 6x6 kernels and the other 2 fully connected layers as convolutional layers with 1x1 kernels. 

{\orange Table~\ref{table:netParams} summarizes the learning parameters used for training and fine-tuning of AlexNet in our experiments. The listed parameters were tuned through an extensive set of trial and error experiments. According to our exploratory experiments, the learning rate and scheduling rate heavily influenced the convergence of CNNs. A learning rate of 0.001 however ensured proper convergence for all 4 applications. A smaller learning rate slowed down convergence and a larger learning rate often caused convergence failures. Our exploratory experiments also indicated that the value of $\gamma $ depended on the speed of convergence. During a fast convergence, the learning rate can be safely decreased after a few epochs, allowing for the use of a small scheduling rate. However, during a slow convergence, a larger scheduling rate is required to maintain a relatively large learning rate. For all 4 applications, we found $\gamma = .95$ to be a reasonable choice. }

\begin{figure}
\centering
\subfloat{\includegraphics[width=1.0\columnwidth]{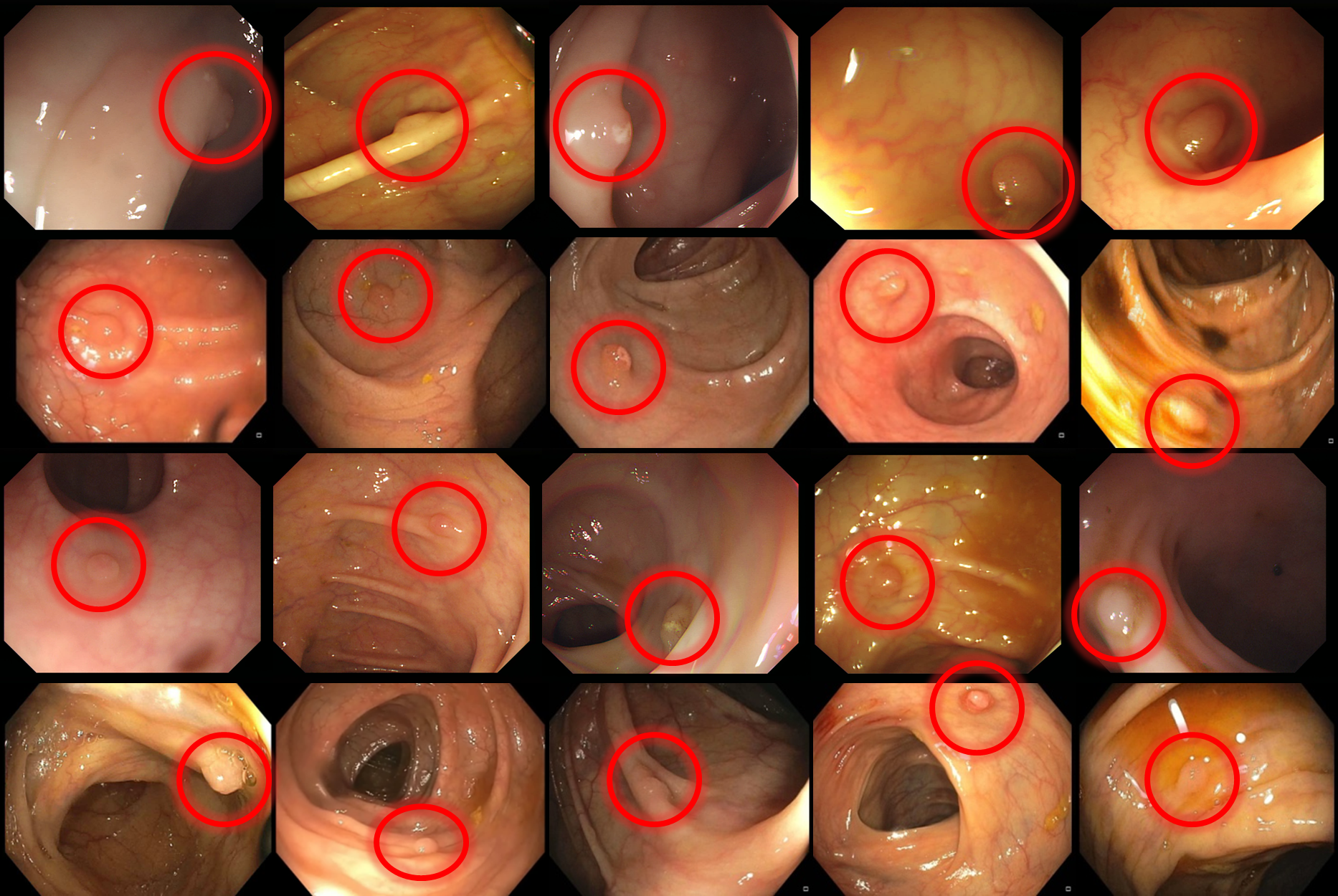}}
\caption{Variations in shape and appearance of polyps in colonoscopy videos.}
\label{fig:PolypEx}
\end{figure}

\begin{figure*}
\centering
\subfloat[]{\includegraphics[width=0.98\columnwidth]{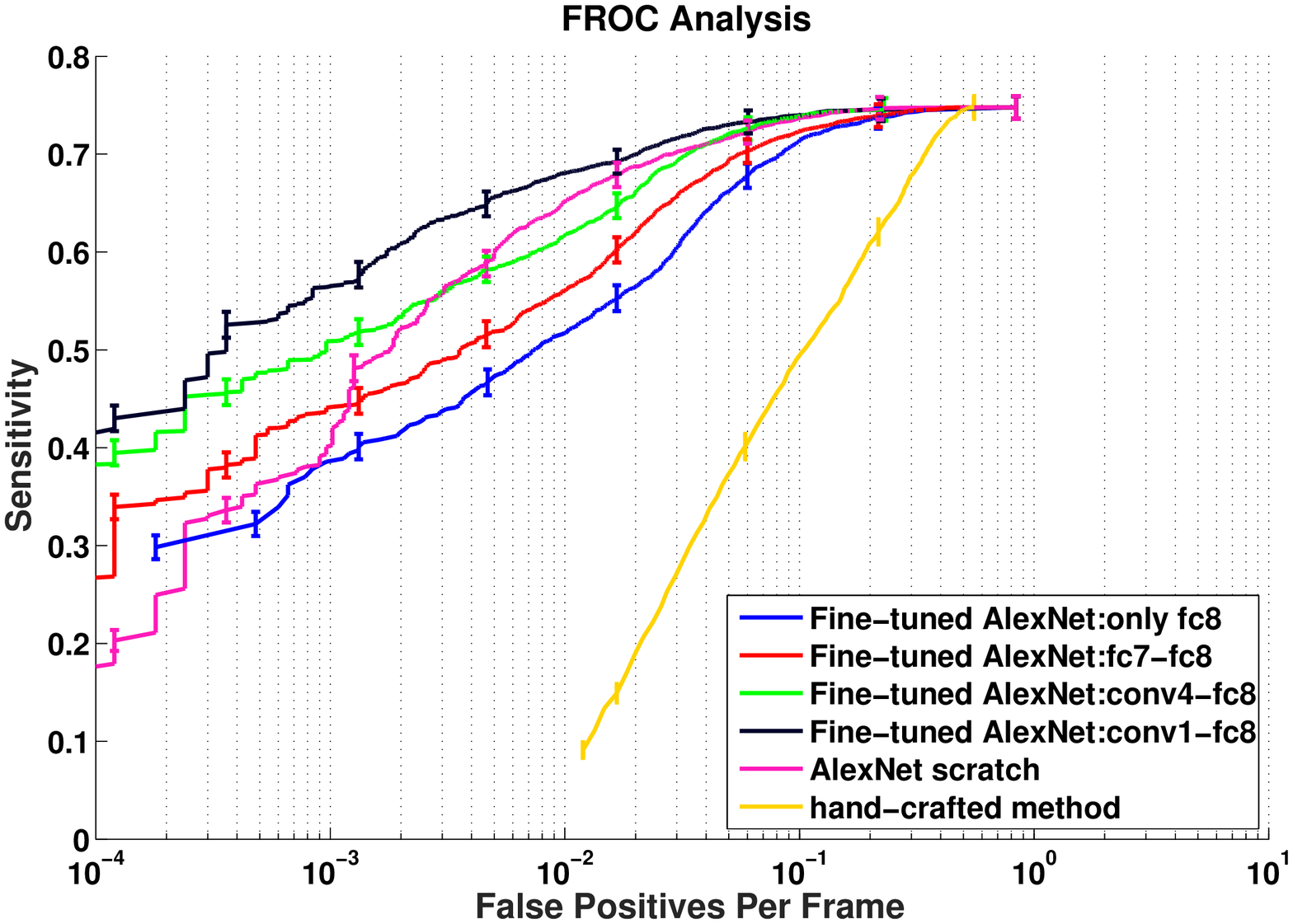}}\hspace{4pt}
\subfloat[]{\includegraphics[width=.98\columnwidth]{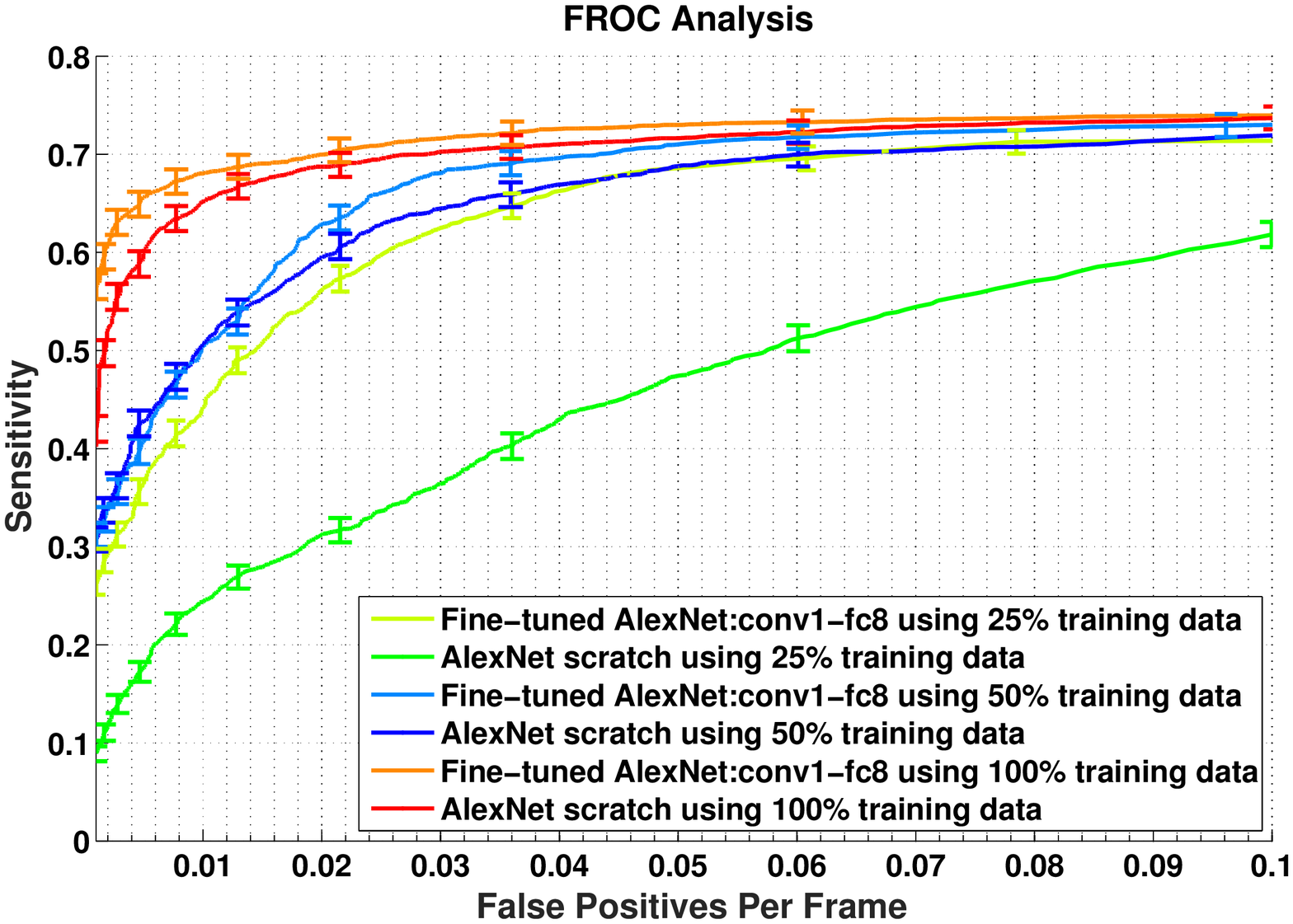}}
\caption{FROC analysis for polyp detection. (a) Comparison between incremental fine-tuning, training from scratch, and a handcrafted approach \cite{ tajbakhsh15d}. (b) Effect of reduction in the training data on the performance of CNNs trained from scratch and deeply fine-tuned CNNs.}
\label{fig:Polyp_FROC}
\end{figure*}

\subsection{Polyp detection}
\label{sec:pd}

{\orange Colonoscopy is the preferred technique for colon cancer screening and prevention. The goal of colonoscopy is to find and remove colonic polyps---precursors to colon cancer. Polyps, as shown in \figurename~\ref{fig:PolypEx}, can appear with substantial variations in color, shape, and size. The challenging appearance of polyps can often lead to misdetection, particularly during long and back-to-back colonoscopy procedures where fatigue negatively affects the performance of colonoscopists. Polyp miss-rates are estimated to be about 4\% to 12\%~\cite{Pabby05,Rijn06, Kim07,Heresbach08}; however, a more recent clinical study~\cite{leufkens12} is suggestive that this misdetection rate may be as high as 25\%. Missed polyps can lead to the late diagnosis of colon cancer with an associated decreased survival rate of less than 10\% for metastatic colon cancer~\cite{Rabeneck03}. Computer-aided polyp detection may enhance optical colonoscopy screening by reducing polyp misdetection.}

{\orange Several computer-aided detection (CAD) systems have been suggested for automatic polyp detection in colonoscopy videos. The early systems \cite{karkanis03, iakovidis05, alexandre08} relied on polyp color and texture for detection. However, limited texture visibility on the surface of polyps and large color variations among polyps hindered the applicability of such systems. More recent systems \cite{Hwang07,Bernal12,Bernal13,Wang13,park12} relied on temporal information and shape information to enhance polyp detection. Shape features proved more effective than color and texture in this regard; however, these features can be misleading without consideration of the context in which the polyp is found.  In our previous works \cite{Nima13,Nima14p,Nima14p2}, culminated in \cite{tajbakhsh15d}, we attempted to overcome the limitation of approaches based solely on polyp shape. Specifically, we suggested a handcrafted approach for combining the shape and context information around the polyp boundaries and demonstrated the superiority of this approach over the other state-of-the-art methods.  }

For training and evaluation, we used our database of 40 short colonoscopy videos. Each colonoscopy frame in our database comes with a binary ground truth image. We randomly divided the colonoscopy videos into a training set containing 3,800 frames with polyps and 15,100 frames without polyps and into a test set containing 5,700 frames with polyps and 13,200 frames without polyps. We applied our handcrafted approach \cite{tajbakhsh15d} to the training and test frames to obtain a set of polyp candidates with the corresponding bounding boxes. At each candidate location, given the available bounding box, we extracted a set of image patches with data augmentation. Specifically, for each candidate, we extracted patches at 3 scales by enlarging the corresponding bounding box by a factor of 1.0x, 1.2x, and 1.5x. At each scale, we extracted patches after we translated the candidate location by 10\% of the resized bounding box in horizontal and vertical directions. We further rotated each resulting patch 8 times by horizontal and vertical mirroring and flipping. We then labeled a patch as positive if the underlying candidate fell inside the ground truth for polyps; otherwise, the candidate was labeled as negative. Because of the relatively large number of negative patches, we collected a stratified set of 100,000 training patches for training and fine-tuning the CNNs. During the test stage, all test patches extracted from a polyp candidate were fed to the trained CNN. We then averaged the probabilistic outputs of the test patches at the candidate level and performed an FROC analysis for performance evaluation.

\figurename~\ref{fig:Polyp_FROC}(a) compares the FROC curve of our handcrafted approach with that of fine-tuned CNNs and a CNN trained from scratch.  {\blue To avoid clutter in the figure, we have shown only a subset of representative FROC curves.} Statistical comparisons between each pair of FROC curves at three operating points are also presented in Table~\ref{table:polyp_statAnal}. {\orange The handcrafted approach is significantly outperformed by all CNN-based scenarios (p$<$.05).} This result is probably because our handcrafted approach used only geometric information to remove false-positive candidates. For fine-tuning, the lowest performance was obtained when only the last layer of AlexNet was updated with colonoscopy data.  {\blue However, fine-tuning the last two layers (FT:fc7-fc8) achieved a significantly higher sensitivity (p$<$.05) at nearly all operating  points compared to the pre-trained AlexNet with only 1 fine-tuned layer (FT:only fc8).  We also observed incremental performance improvement when we included more convolutional layers in the fine-tuning process. Specifically,  the pre-trained CNN with shallow fine-tuning (FT:fc7-fc8) was significantly outperformed by the pre-trained CNNs with a moderate level of fine-tuning (FT:conv5,4,3-fc8) at most of the  operating points. Furthermore, the deeply-tuned CNNs (FT:conv1,2-fc8) achieved a significantly higher sensitivity than the pre-trained CNNs with a moderate level of fine-tuning particularly at low false positive rates.} {\orange Also, as seen in \figurename~\ref{fig:Polyp_FROC}(a), fine-tuning the last few convolutional layers was sufficient to outperform an AlexNet model trained from scratch in a low false positive setting. }

The performance gap between fully trained AlexNet model and their deeply fine-tuned counterparts becomes more evident when fewer training samples are used for training and tuning. To demonstrate this effect, we trained a CNN from scratch and fine-tuned the entire AlexNet using 50\% and 25\% of the entire training samples. We reduced training data at the video level to exclude a fraction of unique polyps from the training set. The results are shown in \figurename~\ref{fig:Polyp_FROC}(b). With a 50\% reduction in training data, a significant performance gap was observed between the CNN trained from scratch and the deeply fine-tuned CNN. With a 25\% reduction in the training data, the fully trained CNN showed dramatic performance degradation, but the deeply fine-tuned CNN still exhibited relatively high performance. These findings strongly favor the use of the fine-tuning approach over full training of a CNN from scratch. 

\begin{figure}
\centering
\subfloat{\includegraphics[width=1.0\columnwidth]{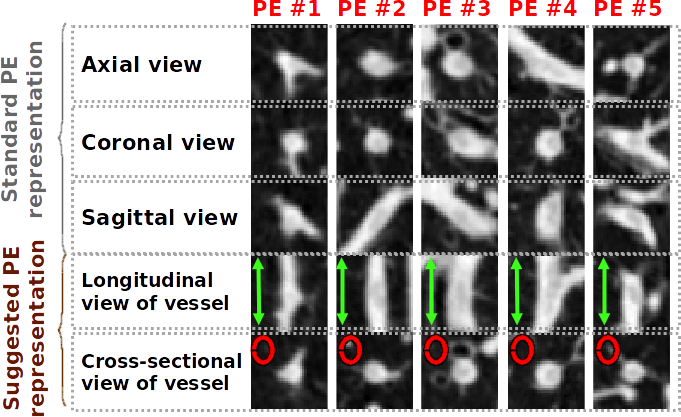}}
\caption{5 different PEs in the standard 3-channel representation and in our suggested 2-channel representation. PEs appear more consistently in our representation. We use our PE representation for the experiments presented herein because it achieves greater classification accuracy and enables improved convergence.}
\label{fig:PEEx}
\end{figure}

\begin{figure*}
\centering
\subfloat[]{\includegraphics[width=.49\linewidth]{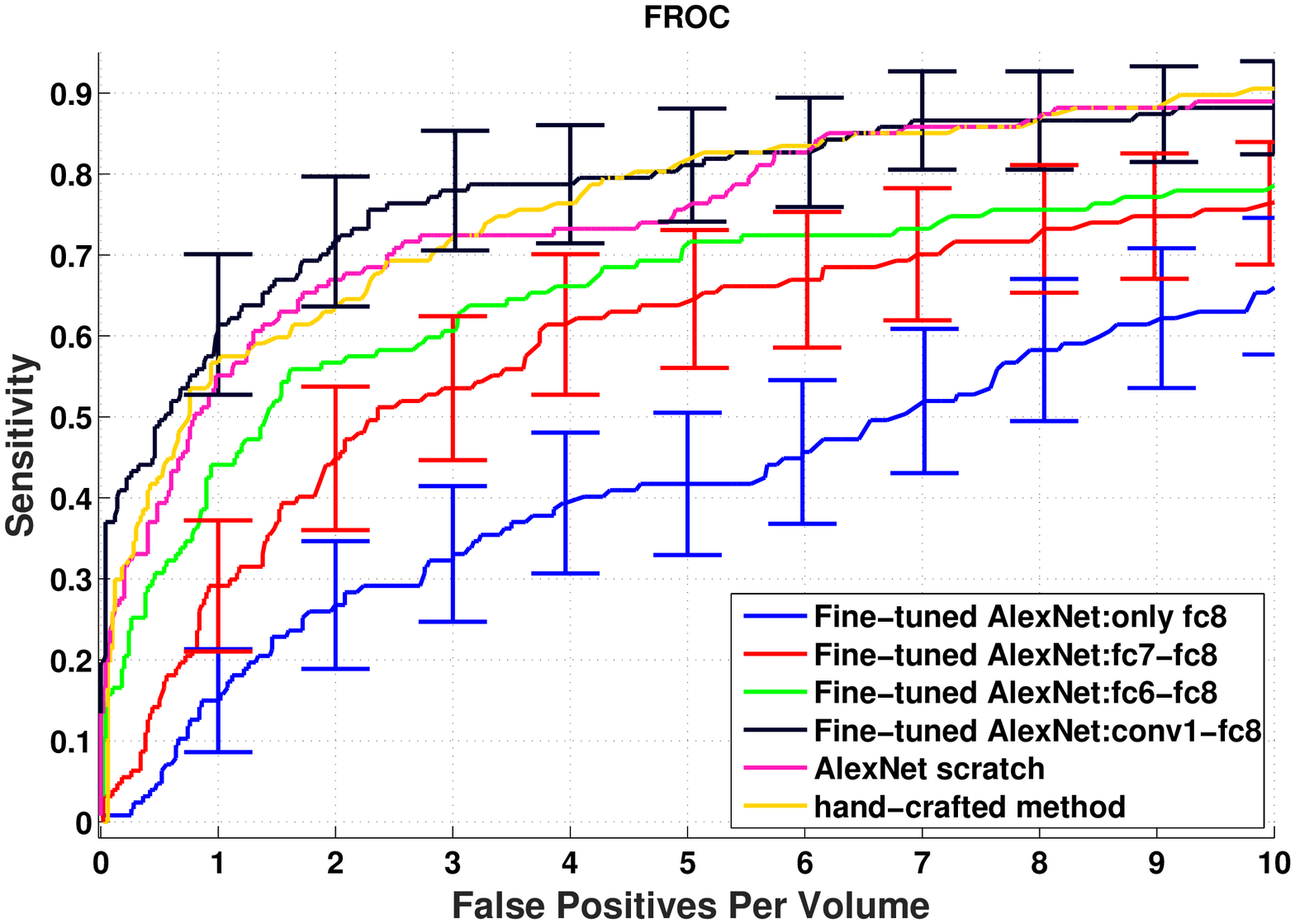}}\hspace{2pt}
\subfloat[]{\includegraphics[width=.49\linewidth]{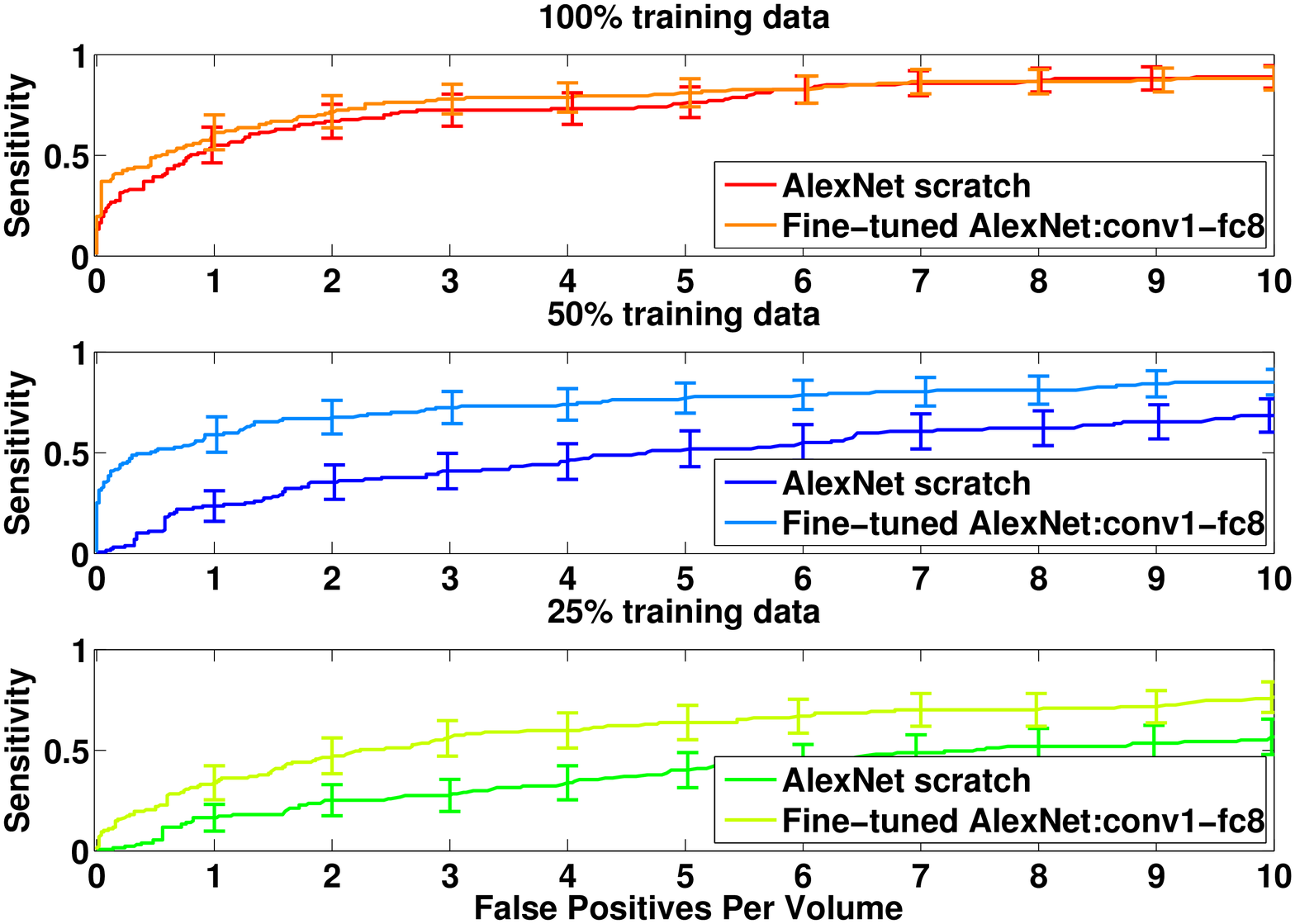}}
\caption{FROC analysis for pulmonary embolism detection. (a) Comparison between incremental fine-tuning, training from scratch, and a handcrafted approach \cite{liang07}. To avoid clutter in the figure, error bars are displayed for only a subset of plots. A more detailed analysis is presented in Table~\ref{table:PE_statAnal}. (b) Effect of reduction in the training data on the performance of CNNs trained from scratch and deeply fine-tuned CNNs.}
\label{fig:PE_FROC}
\end{figure*}

\subsection{Pulmonary embolism detection}
\label{sec:ped}

{\orange A PE is a blood clot that travels from a lower extremity source to the lung, where it causes blockage of the pulmonary arteries. The mortality rate of untreated PE may approach 30\%~\cite{calderthe2005}, but it decreases to as low as 2\% with early diagnosis and appropriate treatment~\cite{sadighchallenges2011}. CT pulmonary angiography (CTPA) is the primary means for PE diagnosis, wherein a radiologist carefully traces each branch of the pulmonary artery for any suspected PEs. CTPA interpretation is a time-consuming task whose accuracy depends on human factors, such as attention span and sensitivity to the visual characteristics of PEs. CAD can have a major role in improving PE diagnosis and decreasing the reading time of CTPA datasets.}

We based our experiments on the PE candidates generated by our previous work \cite{liang07} and the image representation that we suggested for PE in our recently published study \cite{tajbakhsh15c}. Our candidate generation method is an improved version of the tobogganing algorithm \cite{fairfield90} that aims to find an embolus as a dark region surrounded by a brighter background. Our image representation consistently results in 2-channel image patches, which capture PEs in cross-sectional and longitudinal views of vessels (see \figurename~\ref{fig:PEEx}). This unique representation dramatically decreases the variability in the appearance of PEs, enabling us to train more accurate CNNs. {\orange However, since the AlexNet architecture receives color images as its input, the 2-channel image patches must be converted to color patches. For this purpose, we simply repeated the second channel and produced 3-channel RGB-like image patches.} The resulting patches were then used for training and fine-tuning an AlexNet. For performance comparison, we used a handcrafted approach \cite{liang07}, which is arguably one of the most, if not the most, accurate PE CAD system. {\orange The handcrafted approach utilizes the same candidate generation method \cite{liang07}, but uses vessel-based features along with Haralick \cite{haralick73} and wavelet-based features for PE characterization, and finally uses a multi-instance classifier for candidate classification.}

For experiments, we used a database consisting of 121 CTPA datasets with a total of 326 PEs. We first applied the tobogganing algorithm to obtain a crude set of PE candidates. This application resulted in 6,255 PE candidates, of which 5,568 were false positives and 687 were true positives. The number of true positives was far larger than the number of PEs because the tobogganing algorithm can cast several candidates for the same PE. We divided the collected candidates at the patient level into a training set with 434 true positives (199 unique PEs) and 3,406 false positives, and a test set with 253 true positives (127 unique PEs) and 2,162 false positives. For training the CNNs, we extracted patches of 3 different physical sizes, resulting in 10 mm-, 15 mm-, and 20 mm-wide patches. We also translated each candidate location along the direction of the affected vessel 3 times, up to 20\% of the physical size of the patches. We further augmented the training dataset by rotating the longitudinal and cross-sectional vessel planes around the vessel axis, resulting in 5 additional variations for each scale and translation. We formed a stratified training set with 81,000 image patches for training and fine-tuning the CNNs. For testing, we performed the same data augmentation for each test candidate and then computed the overall PE probability by averaging the probabilistic scores generated for the data-augmented patches for each PE candidate.

For evaluation, we performed an FROC analysis by changing a threshold on the probabilistic scores generated for the test PE candidates. \figurename~\ref{fig:PE_FROC}(a) shows the FROC curves for the handcrafted approach, a deep CNN trained from scratch, and {\blue a subset of representative pre-trained CNNs that are fine-tuned in a layer-wise manner.} We have further summarized statistical comparisons between each pair of FROC curves in Table~\ref{table:PE_statAnal}. {\blue  As shown, the pre-trained CNN with two fine-tuned layers (FT:fc7-fc8) achieved a significantly higher sensitivity (p$<$0.05) than that of the pre-trained CNN with only one fine-tuned layer (FT:only fc8). The improved sensitivity was observed at most of the operating points.  However, inclusion of each new layer in the fine-tuning process resulted in only marginal performance improvement, even though the accumulation of such marginal improvements yielded a substantial margin between the deeply fine-tuned CNNs and those with 1, 2, or 3 fine-tuned layers. Specifically, the deeply fine-tuned CNN (FT:conv1-fc8) yielded significantly higher sensitivity (p$<$0.05) than that of the pre-trained CNN with 2 fine-tuned layers (FT:fc7-fc8) at the majority of the operating points shown in \figurename~\ref{fig:PE_FROC}(a). At 3 false positives per volume, the deeply fine-tuned CNN also achieved significantly higher sensitivity (p$<$0.05)  than that of  the pre-trained CNN with three fine-tuned layers (FT:fc7-fc8).} {\orange From \figurename~\ref{fig:PE_FROC}(a), it is also evident that the deeply fine-tuned CNN yielded a non-significant performance improvement over the handcrafted approach. This is probably because the handcrafted approach is an accurate system whose underlying features are specifically and incrementally designed to remove certain types of false detections. Yet, we find it interesting that an end-to-end learning machine can learn such a sophisticated set of features with minimal engineering effort. From \figurename~\ref{fig:PE_FROC}(a), we also observed that the deeply fine-tuned CNN performs on a par with the CNN trained from scratch. }

We further analyzed how the size of training samples influences the competitive performance between the CNN trained from scratch and the deeply fine-tuned CNN. For this purpose, we reduced the training samples at the PE-level to 50\% and 25\%. The results are shown in \figurename~\ref{fig:PE_FROC}(b). With a 50\% reduction in training data, a significant performance gap was observed between the CNN trained from scratch and the deeply tuned CNN in all the operating points. With a 25\% reduction in the training data, we observed a decrease in the overall performance of both CNNs with a smaller yet significant gap between the two curves in most of the operating points. These findings not only favor the use of a deeply fine-tuned CNN but also underscore the importance of large training sets for effective training and fine-tuning of CNNs.

\begin{figure}
\centering
\subfloat{\includegraphics[width=1.0\columnwidth]{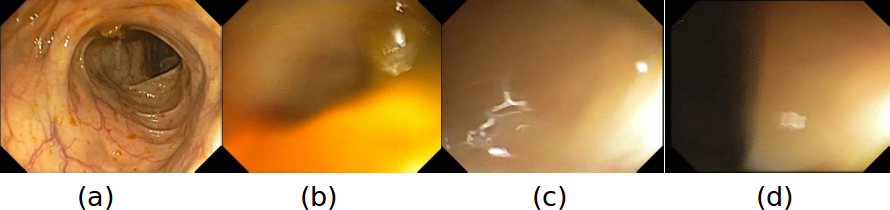}}
\caption{(a) An informative colonoscopy frame. (b,c,d) Examples of non-informative colonoscopy images. The non-informative frames are usually captured during the rapid motion of the scope or during wall contact.}
\label{fig:FramEx}
\end{figure}

%\clearpage

\subsection{Colonoscopy frame classification}
\label{sec:fc}

\begin{figure*}
\centering
\subfloat[]{\includegraphics[width=1.0\columnwidth]{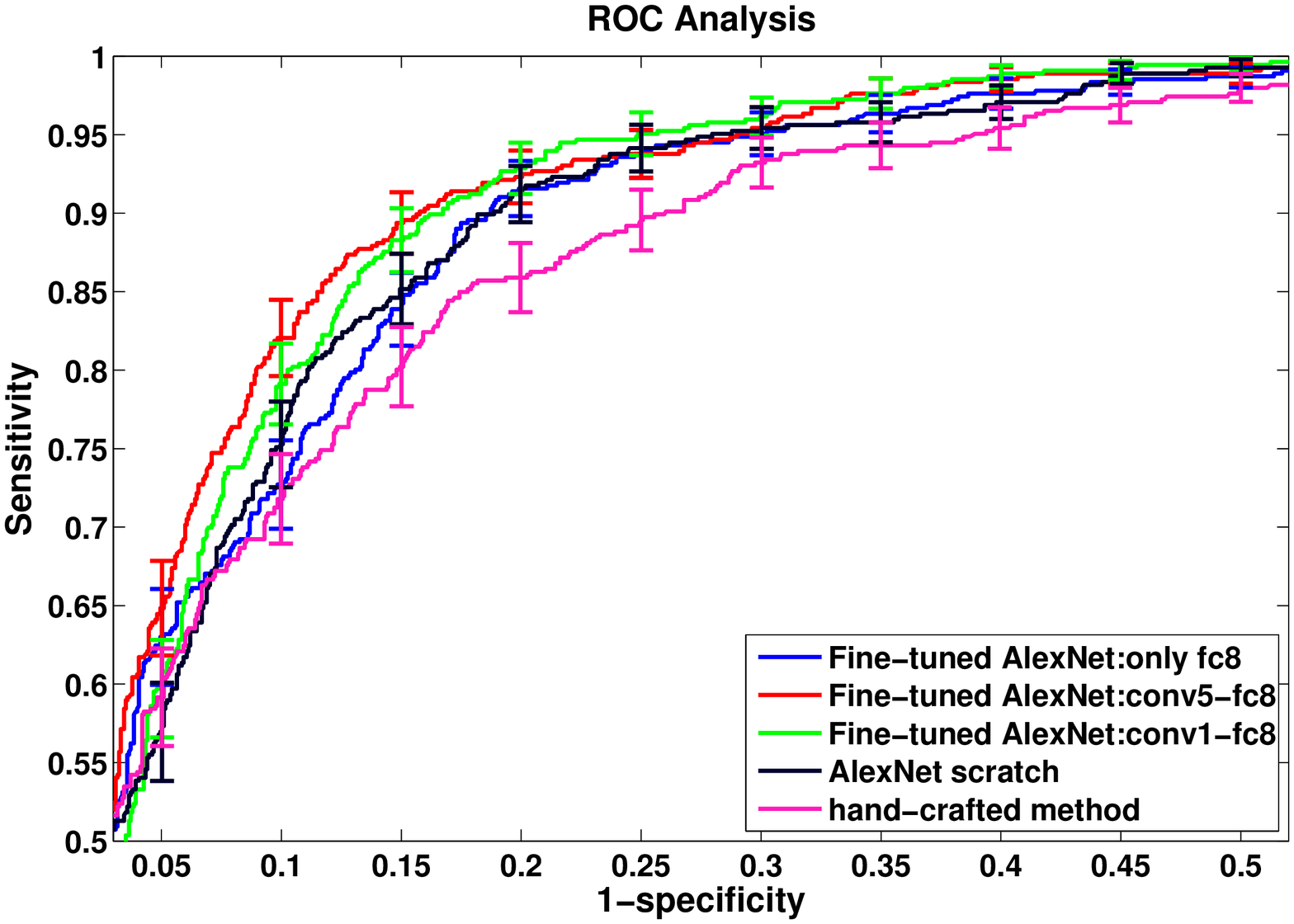}}\hspace{2pt}
\subfloat[]{\includegraphics[width=1.0\columnwidth]{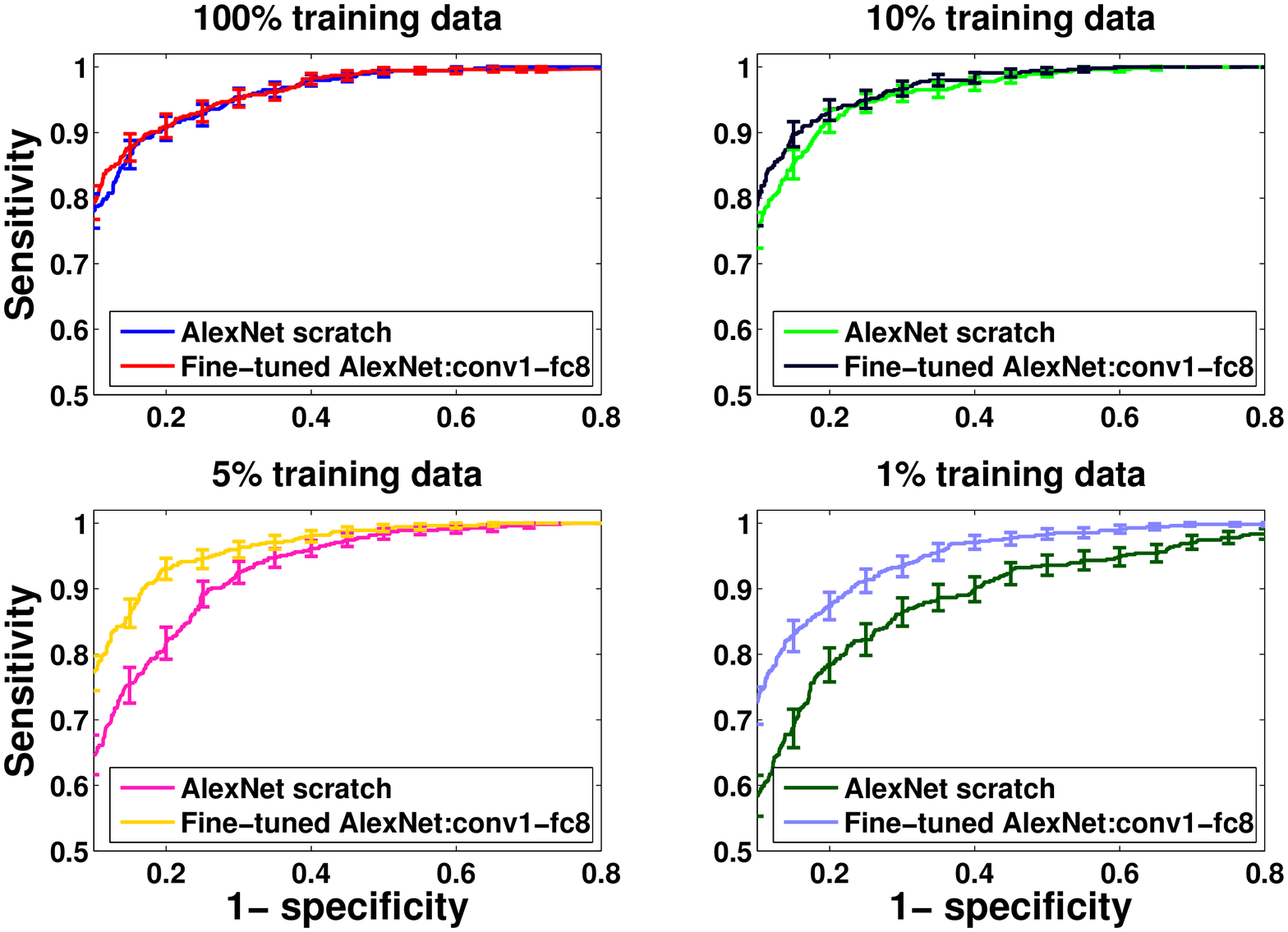}}
\caption{ROC analysis for image quality assessment. (a) Comparison between incremental fine-tuning, training from scratch, and a handcrafted approach \cite{tajbakhsh14q}. (b) Effect of reduction in the training data on the performance of convolutional neural networks (CNNs) trained from scratch vs deeply fine-tuned CNNs.}
\label{fig:QA_ROC}
\end{figure*}

{\orange Image quality assessment can have a major role in objective quality assessment of colonoscopy procedures. Typically, a colonoscopy video contains a large number of non-informative images with poor colon visualization that are not suitable for inspecting the colon or performing therapeutic actions. The larger the fraction of non-informative images in a video, the lower the quality of colon visualization, and thus the lower the quality of colonoscopy. Therefore, one way to measure the quality of a colonoscopy procedure is to monitor the quality of the captured images. Such quality assessment can be used during live procedures to limit low-quality examinations or in a post-processing setting for quality monitoring purposes. }

Technically, image quality assessment at colonoscopy can be viewed as an image classification task whereby an input image is labeled as either \textit{informative} or \textit{non-informative}. \figurename~\ref{fig:FramEx} shows examples of non-informative and informative colonoscopy frames. In our previous work \cite{tajbakhsh14q}, we suggested a handcrafted approach based on local and global features that were pooled from the image reconstruction error. We showed that our handcrafted approach outperformed the other major methods \cite{arnold09,oh07} for quality assessment in colonoscopy videos. In the current effort, we explored the use of deep CNNs as an alternative to a carefully engineered method. Specifically, we compared the performance of our handcrafted approach with that of a deep CNN trained from scratch and a pre-trained CNN that was fine-tuned using the labeled colonoscopy frames in a layer-wise manner..

For experiments, we used 6 complete colonoscopy videos. Considering the expenses associated with annotation of all video frames, we instead sampled each colonoscopy video by selecting 1 frame from every 5 seconds of each video and thereby removed many similar colonoscopy frames. The resulting set was further refined to create a balanced dataset of 4,000 colonoscopy images in which both informative and non-informative classes were represented equally. A trained expert then manually labeled the collected images as informative or non-informative. A gastroenterologist further reviewed the labeled images for corrections. We divided the labeled frames at the video-level into training and test sets, each containing approximately 2,000 colonoscopy frames. For data augmentation, we extracted 200 sub-images of size 227x227 pixels from random locations in each 500x350 colonoscopy frame, resulting in a stratified training set with approximately 40,000 sub-images. During the test stage, the probability of each frame being informative was computed as the average probabilities assigned to its randomly cropped sub-images.

We used an ROC analysis for performance comparisons between the CNN-based scenarios and  handcrafted approach. The results are shown in \figurename~\ref{fig:QA_ROC}(a). {\blue To avoid clutter in the figure, we have shown only a subset of representative ROC curves. We have, however, summarized the statistical comparisons between all ROC curves at 10\%, 15\%, and 20\% false positive rates in Table~\ref{table:QA_statAnal}. We observed that all CNN-based scenarios significantly outperformed the handcrafted approach in at least one of the above 3  operating points. We also observed that fine-tuning the pre-trained CNN halfway through the network (FT:conv4-fc8 and FT:conv5-fc8) not only significantly outperformed shallow-tuning but also was superior to a deeply fine-tuned CNN (FT:conv1-fc8) at 10\% and 15\% false positive rates. This was probably because the kernels learned in the early layers of the CNN were suitable for image quality assessment and thus their fine-tuning was unnecessary. Furthermore, while the CNN trained from scratch outperformed the pre-trained CNN with shallow fine-tuning (FT:only fc8), it was outperformed by the pre-trained CNN with a moderate level of fine-tuning (FT:conv5-fc8). Therefore, the fine-tuning scheme was superior to the full training scheme from scratch.}

To examine how the performance of CNNs changes with respect to the size of the training data, we decreased the number of training samples by factors of 1/10, 1/20, and 1/100. Comparing these with other applications, we considered a further reduction in the size of the training dataset because a moderate decrease did not influence the performance of CNNs substantially. As shown in \figurename~\ref{fig:QA_ROC}(b), both deeply fine-tuned CNNs and fully trained CNN showed insignificant performance degradation even when using 10\% of the original training set. However, further reduction in the size of the training set substantially degraded the performance of fully trained CNNs and, to a largely less extent, the performance of deeply fine-tuned CNNs. The relatively high performance of the deeply fine-tuned CNNs, even with a limited training set, indicates the usefulness of the kernels learned from ImageNet for colonoscopy frame classification.

\begin{figure}
\centering
\subfloat{\includegraphics[width=1.0\columnwidth]{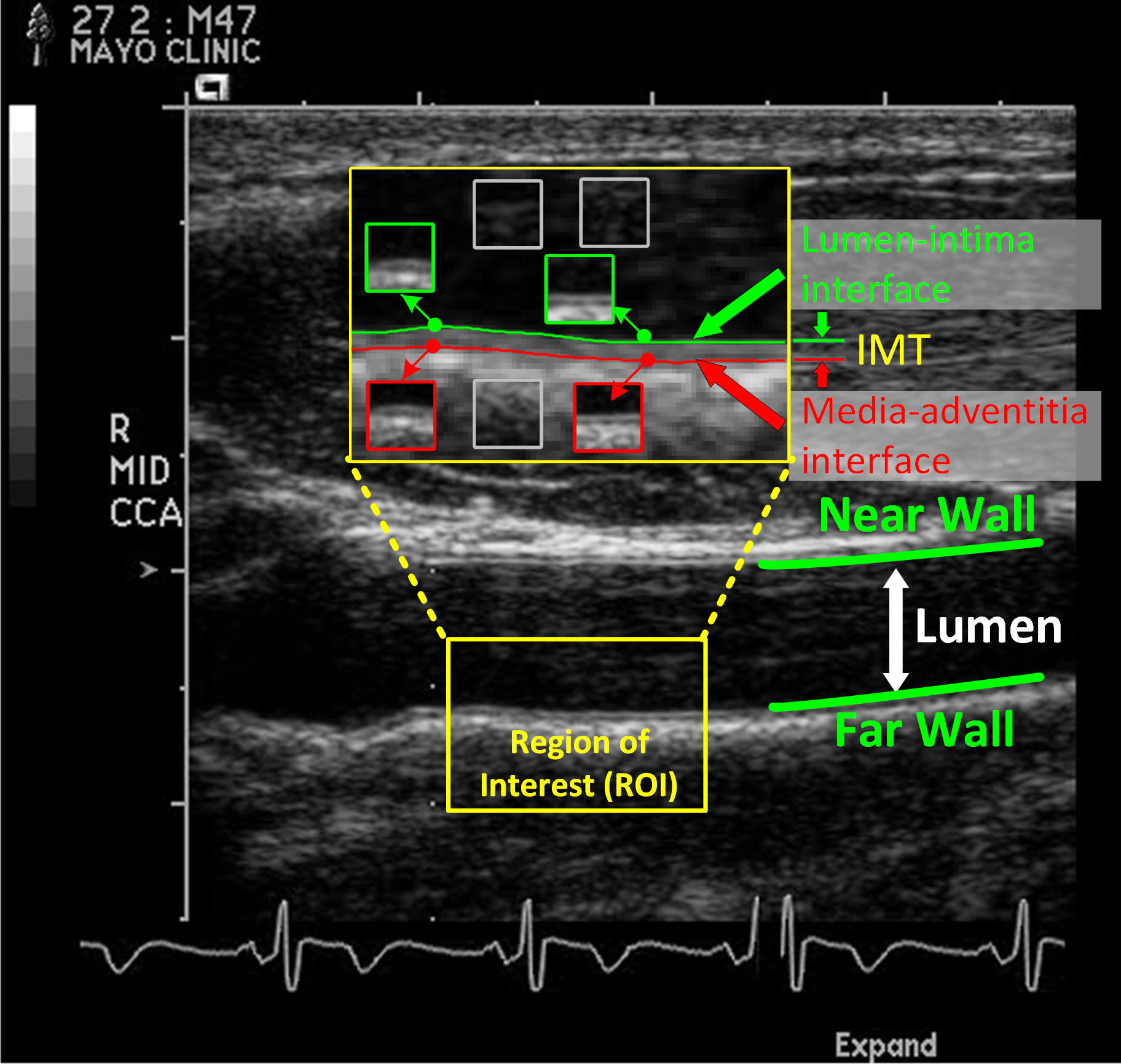}}
\caption{Intima media thickness (IMT) is measured within a region of interest after the lumen-intima and media-adventitia interfaces are segmented. For automatic interface segmentation, we trained a 3-way convolutional neural network whose training patches were extracted from each of these interfaces (highlighted in red and green) and far from the interfaces (highlighted in gray).}
\label{fig:CIMT_intro}
\end{figure}

\begin{figure*}
\centering
\subfloat{\includegraphics[width=0.98\linewidth]{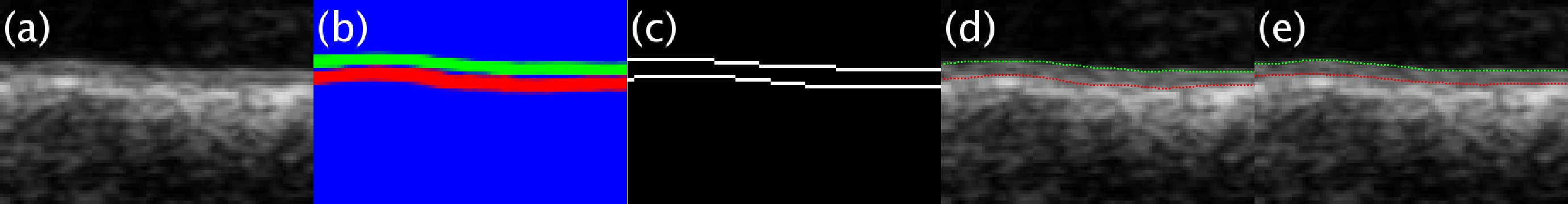}}
\caption{The test stage of lumen-intima and media-adventitia interface segmentation. (a) A test region of interest. (b) The corresponding confidence map generated by the convolutional neural network. The green and red colors indicate the likelihood of a lumen-intima interface and media-adventitia interface, respectively. (c) The thick probability band around each interface is thinned by selecting the largest probability for each interface in each column. (d) The step-like boundaries are smoothed using 2 open snakes. {\orange (e) Interface segmentation from the ground truth.}}
\label{fig:CIMT_Seg}
\end{figure*}

\subsection{Intima-media boundary segmentation}
\label{sec:cimt}
Carotid intima-media thickness (CIMT), a noninvasive ultrasonography method, has proven valuable for cardiovascular risk stratification. The CIMT is defined as the distance between the lumen-intima and media-adventitia interfaces at the far wall of the carotid artery (\figurename~\ref{fig:CIMT_intro}). The CIMT measurement is performed by manually tracing the lumen-intima and media-adventitia interfaces in a region of interest (ROI), followed by calculation of the average distance between the traced interfaces. However, manual tracing of the interfaces is time-consuming and tedious. {\orange Therefore, several methods \cite{menchon15,menchon13,petroudi12,xu12} have been developed to allow automatic CIMT image interpretation. The suggested methods are more or less based on handcrafted techniques whose performance may vary according to image quality and the level of artifacts present within the images. }

\begin{figure*}
\centering
\subfloat[]{\includegraphics[width=0.48\linewidth]{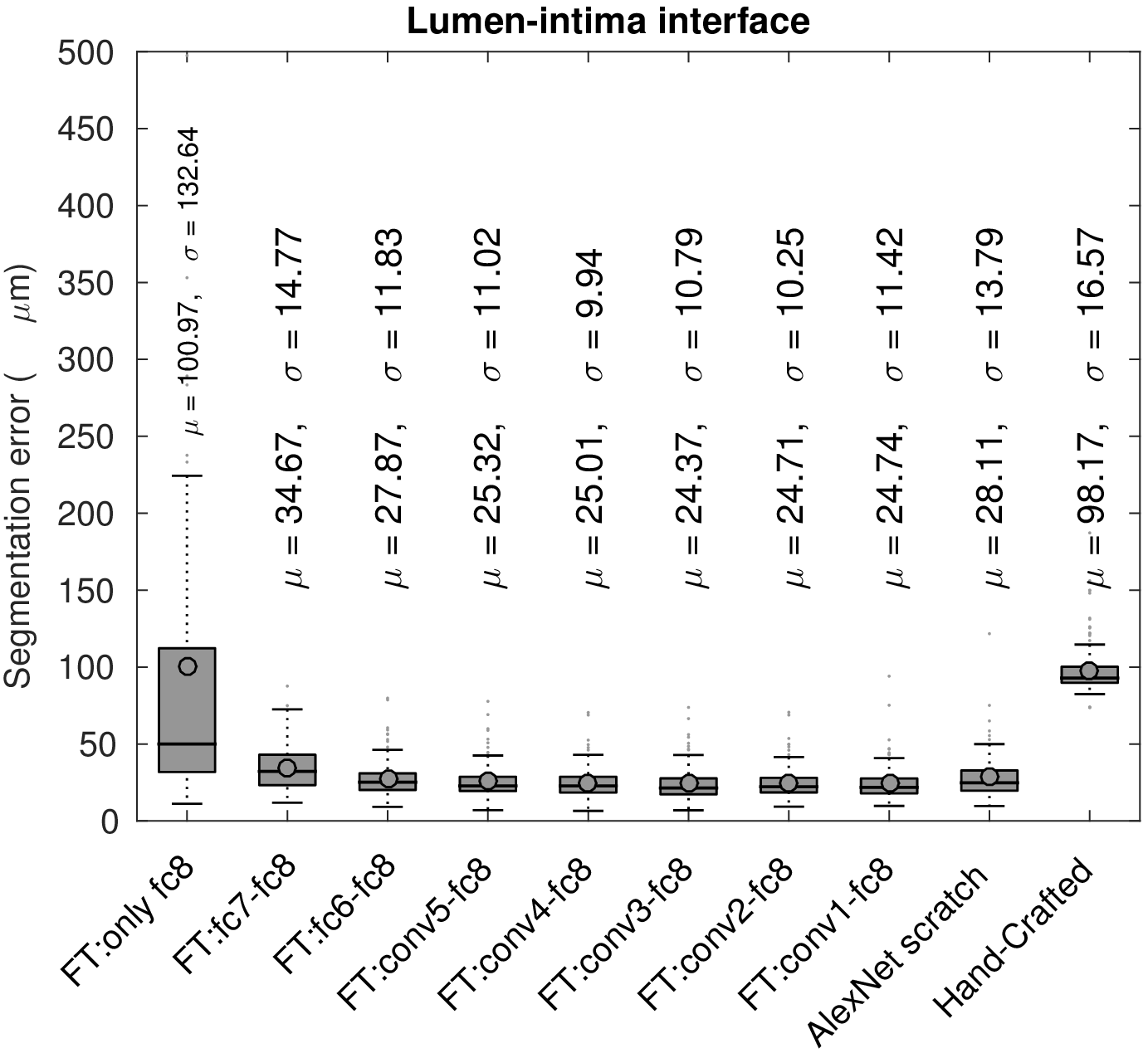}}\hspace{4pt}
\subfloat[]{\includegraphics[width=0.48\linewidth]{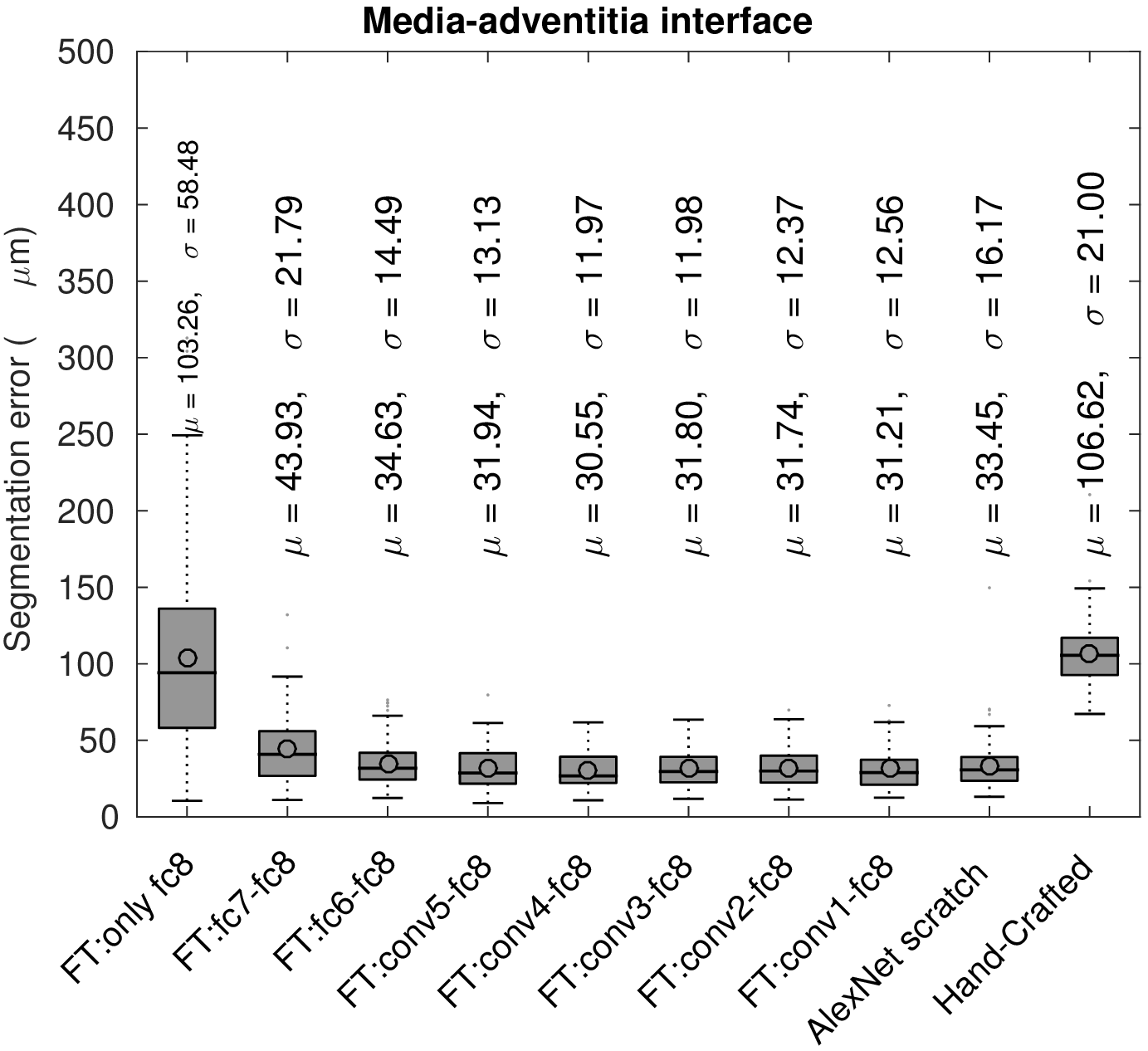}}
\caption{Box plots of segmentation error for (a) the lumen-intima interface and (b) the media-adventitia interface.}
\label{fig:CIMT_boxplot}
\end{figure*}

We formulated this interface segmentation task as a 3-class classification problem wherein the goal was to classify every pixel in the ROI into 3 categories: a pixel on the lumen-intima interface, a pixel on the media-adventitia interface, or a non-interface pixel. For this classification problem, we trained a 3-way CNN using the training patches collected from the lumen-intima interface and media-adventitia interface, as well as from other random locations far from the desired interfaces.  \figurename~\ref{fig:CIMT_intro} illustrates how these patches are extracted from an ultrasonography frame.

\figurename~\ref{fig:CIMT_Seg} shows how a CNN-based system traces the interfaces for a given test ROI. The trained CNN is first applied to each pixel within the test ROI in a convolutional manner, generating 2 confidence maps of the same size as the ROI, with the first map showing the probability of a pixel residing on the lumen-intima interface and the second map showing the probability of a pixel residing on the media-adventitia interface. For visualization convenience, we merged these 2 confidence maps into 1 color-coded confidence map in which the green and red colors indicate the likelihood of being a lumen-intima interface and a media-adventitia interface, respectively. As shown in \figurename~\ref{fig:CIMT_Seg}(b), the probability band of each interface is too thick to accurately measure intima-media thickness. To resolve this issue, we obtained thinner interfaces by scanning the confidence map column by column to search for rows with the maximum response for each of the 2 interfaces, yielding a 1-pixel boundary with a step-like shape around each interface, as shown in \figurename~\ref{fig:CIMT_Seg}(c). To smooth the boundaries, we used 2 active contour models (snakes) \cite{liang06}, one for the lumen-intima interface and one for the media-adventitia interface. The open snakes were initialized with the current step-like boundaries and then kept deforming until they took the actual shapes of the interfaces. \figurename~\ref{fig:CIMT_Seg}(d) shows the converged snakes for the test ROI. We computed intima-media thickness as the average of the vertical distances between the 2 open snakes.

For the experiments, we used a database of 92 CIMT videos. The expert reviews each video to determine 3 ROIs for which the CIMT can be measured reliably. To create the ground truth, lumen-intima and media-adventitia interfaces were annotated as the consensus of 2 experts for each of the 276 ROIs. We divided the ROIs at the subject-level into a training set with 144 ROIs and a test set with 132 ROIs. For training and fine-tuning the CNNs, we extracted a stratified set of 200,000 training patches from the training ROIs. {\orange Because the AlexNet architecture used in our study required color patches as its input, each extracted gray-scale patch was converted to a color patch by repeating the gray channel thrice.} Note that we did not perform data augmentation for the positive patches, for 2 reasons. First, 92x60 ROIs allow us to collect a large number of patches around the lumen-intima and media-adventitia interfaces, eliminating the need for any further data augmentation. Second, given the relatively small distance between the 2 interfaces, translation-based data augmentation would inject a large amount of label noise, which would negatively affect the convergence and the overall performance of the CNNs. In the test stage, we measured the error of interface segmentation as the average distance between the expert-annotated interfaces and those produced by the systems. For a more detailed analysis, we measured segmentation error for the lumen-intima and media-adventitia interfaces separately.

\figurename~\ref{fig:CIMT_boxplot} shows the box plots of segmentation error for each interface. The whiskers were plotted according to Tukey method. {\blue For easier quantitative comparisons, we have also shown the average and standard deviation of the localization error above each boxplot.}  The segmentation error for the media-adventitia interface was generally greater than the lumen-intima interface, which was expected because of the relatively more challenging image characteristics of the media-adventitia interface. {\orange For both interfaces, holding all the layers fixed except the last layer (FT: only fc8) resulted in the lowest performance, which was comparable to that of the handcrafted approach \cite{sharma14}.  However, inclusion of layer fc7 in the fine-tuning process (FT:fc7-fc8) led to a significant decrease (p$<$.0001) in  segmentation error for both interfaces. {\blue The reduced localization error was also significantly lower (p$<$ .0001) than that of  the handcrafted approach. We observed another significant drop (p$<$.001) in the localization error of both interfaces after fine-tuning layer fc6; however, this error was still significantly larger  (p$<$.001) than that of the deeply fine-tuned AlexNet (FT:conv1-fc8). We observed a localization error comparable to that of the deeply fine-tuned AlexNet only after inclusion of layer conv5 in the fine-tuning process. With deeper fine-tuning, we obtained only marginal decrease in the localization error for both interfaces.}  Furthermore, the  localization error obtained by the deeply fine-tuned CNN was significantly lower than that of the CNN trained from scratch for media-adventitia interface (p$<$.05) and for Lumen-intima interface (p$<$.0001), indicating the superiority of the fine-tuning scheme over the training scheme from scratch.}   Of note, we observed no significant performance degradation for either deeply fine-tuned CNNs or fully trained CNNs, even after reducing the training patches to a single patient. This outcome resulted because each patient in our database provided approximately 12 ROIs, which enabled the extraction of a large number of distinct training patches that could be used for training and for fine-tuning the deep CNNs.

\section{Discussion}

In this study, to ensure generalizability of our findings, we considered 4 common medical imaging problems from 3 different imaging modality systems. Specifically, we chose PE detection as representative of computer-aided lesion detection in 3-dimensional volumetric images, polyp detection as representative of computer-aided lesion detection in 2-dimensional images, intima-media boundary segmentation as representative of machine learning-based medical image segmentation, and colonoscopy image quality assessment as representative of medical image classification. These applications differ because they require solving problems at different image scales. For instance, although intima-media boundary segmentation and PE detection may require the examination of a small sub-region within the images, polyp detection and frame classification demand far larger receptive fields. Therefore, we believe that the chosen applications encompass a variety of applications relevant to the field of medical imaging.

We thoroughly investigated the potential for fine-tuned CNNs in the context of medical image analysis as an alternative to training deep CNNs from scratch. We performed our analyses using both large training sets and reduced training sets. When using complete datasets, we observed that shallow tuning of the pre-trained CNNs most often led to a performance inferior to CNNs trained from scratch, whereas with deeper fine-tuning, we obtained performance comparable and even superior to CNNs trained from scratch. The performance gap between deeply fine-tuned CNNs and those trained from scratch widened when the size of training sets was reduced, which led us to conclude that fine-tuned CNNs should always be the preferred option regardless of the size of training sets available. 

\begin{figure*}
\centering
\subfloat{\includegraphics[width=1.0\linewidth]{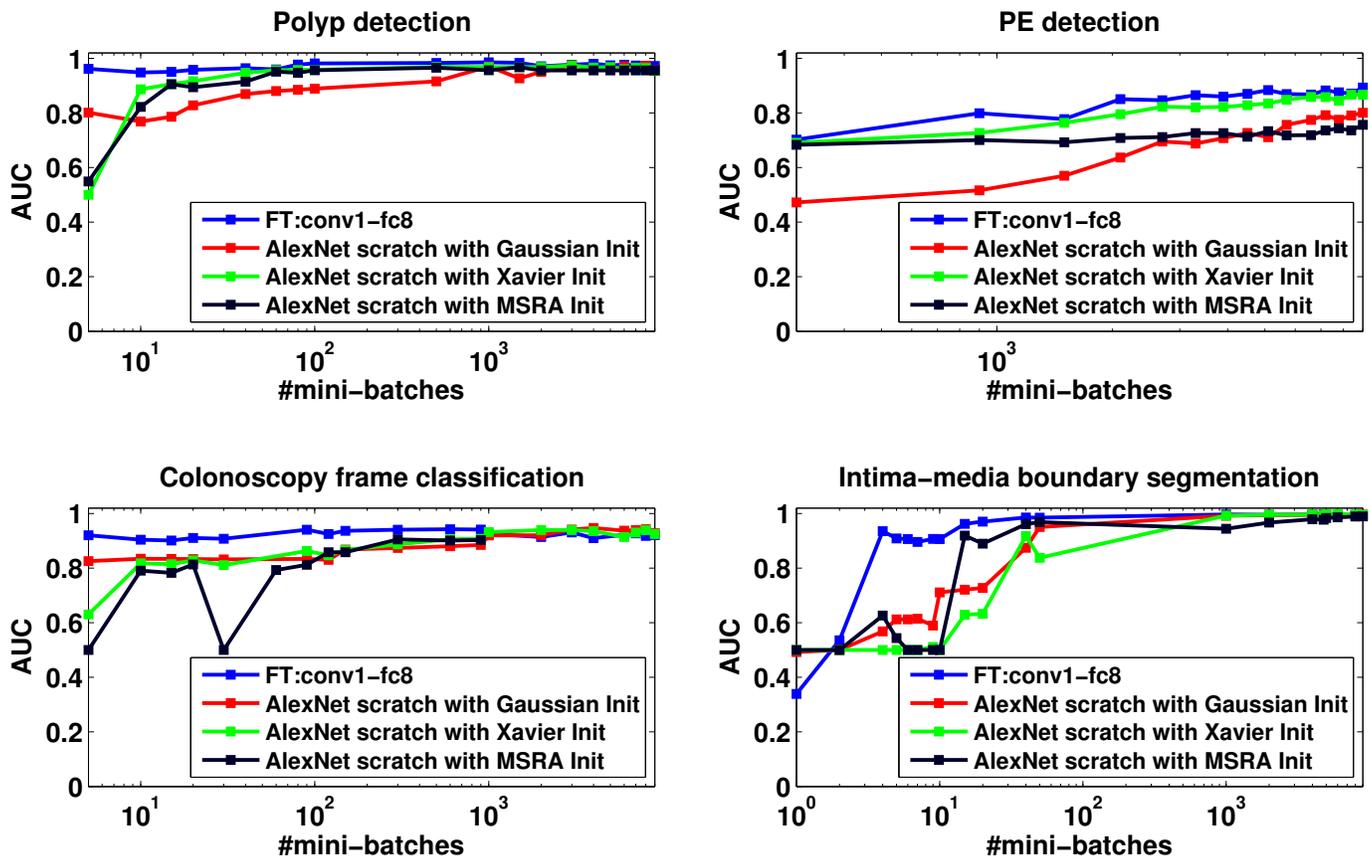}}
\caption{{\orange Convergence speed for a deeply fine-tuned CNN and CNNs trained from scratch with three different initialization techniques.}}
\label{fig:trSpeed}
\end{figure*}

{\orange Another advantage of fine-tuned CNNs is the speed of convergence. To demonstrate this advantage, we compare the speed of convergence  for  a deeply fine-tuned CNN and a CNN trained from scratch in \figurename~\ref{fig:trSpeed}. For a thorough comparison, we used  3 different techniques to initialize the weights of the fully trained CNNs: 1) a method commonly known as Xavier, which was suggested in \cite{glorot10}, 2) a revised version of Xavier called MSRA, which was suggested in \cite{he15}, and a basic random initialization method based on Gaussian distributions. In this analysis, we computed the AUC on the validation data as a measure of convergence. Specifically, each snapshot of the model was applied to the patches of the validation set and then the classification performance was evaluated using an ROC analysis. Because we dealt with a 3-class classification problem for the ask of intimia-media boundary segmentation, we merged the 2 interface classes into a positive class and then computed the AUC for the resulting binary classification (interface vs. background).  As shown, the fine-tuned CNN quickly reaches its maximum performance, but the CNNs trained from scratch require longer training in order to reach their highest performance.  Furthermore, the use of different initialization techniques led to different trends of convergence, even though we observed no significant performance gain after complete convergence {\blue except for PE detection.}}

We observed that the depth of fine-tuning is fundamental to achieving accurate image classifiers. Although shallow tuning or updating the last few convolutional layers is sufficient for many applications in the field of computer vision to achieve state-of-the-art performance, we discovered that a deeper level of tuning is essential for medical imaging applications. For instance, we observed a marked performance gain using deeply fine-tuned CNNs, particularly for polyp detection and intima-media boundary segmentation, probably because of the substantial difference between these applications and the database with which the pre-trained CNN was constructed. However, we did not observe a similarly profound performance gain for colonoscopy frame classification, which we attribute to the relative similarity between ImageNet and the colonoscopy frames in our database. Specifically, both databases use high-resolution images with similar low-level image information, which is why fine-tuning the late convolutional layers, which have application-specific features, is sufficient to achieve high-level performance for colonoscopy frame classification.

We based our experiments on the AlexNet architecture because a pre-trained AlexNet model was available in the Caffe library and that this architecture was deep enough that we could investigate the impact of the depth of fine-tuning on the performance of pre-trained CNNs. Alternatively, deeper architectures---such as VGGNet and GoogleNet---could have been used. Deeper architectures have recently shown relatively high performance for challenging computer vision tasks, but we do not anticipate a significant performance gain through the use of deeper architectures for medical imaging applications. We emphasize that the objective of this work was not to achieve the highest performance for a number of different medical imaging tasks but to examine the capabilities of fine-tuning in comparison with the training scheme from scratch. For these purposes, AlexNet is a reasonable architectural choice.

{\blue We would like to acknowledge that the performance curves reported for different models and applications may not be the best that we could achieve for each experiment. This sub-optimal performance is related to the choice of the hyper-parameters of CNNs that can influence  the speed of convergence and final accuracy of a model. Although we attempted to find the working values of these parameters, finding the optimal values was not feasible given the large number of CNNs studied in our paper and that training each CNN was a time-consuming process even on the high-end GPUs. Nevertheless, this issue may not change our overall conclusions as the majority of the CNNs used in our comparisons are pre-trained models that may be less affected by the choice of hyper-parameters than the CNNs trained from scratch.}

{\orange In this study, due to space constraints, we were not able to cover all medical imaging modalities. For instance, we did not study the performance of fine-tuning in MR images or histopathology images, for which full training of CNNs from scratch had shown promising performance. However, considering the successful knowledge transfer from natural images to CT, ultrasound, and endoscopy applications, we surmise that fine-tuning would succeed in other medical applications as well. Furthermore, our study was focused on fine-tuning of a pre-trained supervised model. However, a pre-trained unsupervised model such as those obtained by restricted Boltzmann machines (RBMs) or convolutional RBMs \cite{lee09} could also be considered, even though the availability of ImageNet database with millions of labeled images from 1000 semantic classes may make the use of a pre-trained supervised model a natural choice for fine-tuning. Nevertheless, unsupervised models are still useful for 1D signal processing due to the absence of a large database of labeled 1D signals. For instance, fine-tuning of an unsupervised model was used in \cite{abdel14} for acoustic speech recognition and in \cite{wulsin11} for detection of epilepsy in EEG recordings.}

 \section{Conclusion}
 
In this paper, we aimed to address the following central question in the context of medical image analysis: \emph{Can the use of pre-trained deep CNNs, with sufficient fine-tuning, eliminate the need for training a deep CNN from scratch?} Our extensive experiments, based on 4 distinct medical imaging applications from 3 different imaging modality systems, have demonstrated that deeply fine-tuned CNNs are useful for medical image analysis, performing as well as fully trained CNNs and even outperforming the latter when limited training data are available. Our results are important because they show that knowledge transfer from natural images to medical images is possible, even though the relatively large difference between source and target databases is suggestive that such application may not be possible. {\orange We also have observed that the required level of fine-tuning differed from one application to another.  Specifically, for PE detection, we achieved performance saturation after fine-tuning the late fully connected layers; for colonoscopy frame classification, we achieved the highest performance through fine-tuning the late and middle layers; and for interface segmentation and polyp detection, we observed the highest performance by  fine-tuning all layers in the pre-trained CNN.} {\blue Our findings suggest that for a particular application, neither shallow tuning nor deep tuning may be the optimal choice. Through the layer-wise fine-tuning, one can learn the effective depth of tuning, as it depends on the application at hand and the amount of labeled data available for tuning. Layer-wise fine-tuning may offer a practical way to achieve the best performance for the application at hand based on the amount of available data.} Our experiments further confirm the potential of CNNs for medical imaging applications because both deeply fine-tuned CNNs and fully trained CNNs outperformed the corresponding handcrafted alternatives.

\bibliographystyle{IEEEtran}
\bibliography{RefJpaper}

\begin{thebibliography}{10}
\providecommand{\url}[1]{#1}
\csname url@samestyle\endcsname
\providecommand{\newblock}{\relax}
\providecommand{\bibinfo}[2]{#2}
\providecommand{\BIBentrySTDinterwordspacing}{\spaceskip=0pt\relax}
\providecommand{\BIBentryALTinterwordstretchfactor}{4}
\providecommand{\BIBentryALTinterwordspacing}{\spaceskip=\fontdimen2\font plus
\BIBentryALTinterwordstretchfactor\fontdimen3\font minus
  \fontdimen4\font\relax}
\providecommand{\BIBforeignlanguage}[2]{{%
\expandafter\ifx\csname l@#1\endcsname\relax
\typeout{** WARNING: IEEEtran.bst: No hyphenation pattern has been}%
\typeout{** loaded for the language `#1'. Using the pattern for}%
\typeout{** the default language instead.}%
\else
\language=\csname l@#1\endcsname
\fi
#2}}
\providecommand{\BIBdecl}{\relax}
\BIBdecl

\bibitem{fukushima80}
K.~Fukushima, ``Neocognitron: A self-organizing neural network model for a
  mechanism of pattern recognition unaffected by shift in position,''
  \emph{Biological cybernetics}, vol.~36, no.~4, pp. 193--202, 1980.

\bibitem{le98}
Y.~LeCun, L.~Bottou, Y.~Bengio, and P.~Haffner, ``Gradient-based learning
  applied to document recognition,'' \emph{Proceedings of the IEEE}, vol.~86,
  no.~11, pp. 2278--2324, 1998.

\bibitem{le15}
Y.~LeCun, Y.~Bengio, and G.~Hinton, ``Deep learning,'' \emph{Nature}, vol. 521,
  no. 7553, pp. 436--444, 2015.

\bibitem{breakthrough}
``Online: available at
  \url{http://www.technologyreview.com/featuredstory/513696/deep-learning/},.''

\bibitem{szegedy14}
C.~Szegedy, W.~Liu, Y.~Jia, P.~Sermanet, S.~Reed, D.~Anguelov, D.~Erhan,
  V.~Vanhoucke, and A.~Rabinovich, ``Going deeper with convolutions,''
  \emph{arXiv preprint arXiv:1409.4842}, 2014.

\bibitem{simonyan14}
K.~Simonyan and A.~Zisserman, ``Very deep convolutional networks for
  large-scale image recognition,'' \emph{arXiv preprint arXiv:1409.1556}, 2014.

\bibitem{zeiler14}
M.~D. Zeiler and R.~Fergus, ``Visualizing and understanding convolutional
  networks,'' in \emph{Computer Vision--ECCV 2014}.\hskip 1em plus 0.5em minus
  0.4em\relax Springer, 2014, pp. 818--833.

\bibitem{Eigen13}
D.~Eigen, J.~Rolfe, R.~Fergus, and Y.~LeCun, ``Understanding deep architectures
  using a recursive convolutional network,'' \emph{arXiv preprint
  arXiv:1312.1847}, 2013.

\bibitem{erhan09}
D.~Erhan, P.-A. Manzagol, Y.~Bengio, S.~Bengio, and P.~Vincent, ``The
  difficulty of training deep architectures and the effect of unsupervised
  pre-training,'' in \emph{International Conference on artificial intelligence
  and statistics}, 2009, pp. 153--160.

\bibitem{razavian14}
A.~S. Razavian, H.~Azizpour, J.~Sullivan, and S.~Carlsson, ``{CNN} features
  off-the-shelf: an astounding baseline for recognition,'' in \emph{Computer
  Vision and Pattern Recognition Workshops (CVPRW), 2014 IEEE Conference
  on}.\hskip 1em plus 0.5em minus 0.4em\relax IEEE, 2014, pp. 512--519.

\bibitem{azizpour14}
H.~Azizpour, A.~S. Razavian, J.~Sullivan, A.~Maki, and S.~Carlsson, ``From
  generic to specific deep representations for visual recognition,''
  \emph{arXiv preprint arXiv:1406.5774}, 2014.

\bibitem{penatti}
O.~Penatti, K.~Nogueira, and J.~Santos, ``Do deep features generalize from
  everyday objects to remote sensing and aerial scenes domains?'' in
  \emph{Proceedings of the IEEE Conference on Computer Vision and Pattern
  Recognition Workshops}, 2015, pp. 44--51.

\bibitem{zhang94}
W.~Zhang, K.~Doi, M.~L. Giger, Y.~Wu, R.~M. Nishikawa, and R.~A. Schmidt,
  ``Computerized detection of clustered microcalcifications in digital
  mammograms using a shift-invariant artificial neural network,'' \emph{Medical
  Physics}, vol.~21, no.~4, pp. 517--524, 1994.

\bibitem{chan95}
H.-P. Chan, S.-C.~B. Lo, B.~Sahiner, K.~L. Lam, and M.~A. Helvie,
  ``Computer-aided detection of mammographic microcalcifications: Pattern
  recognition with an artificial neural network,'' \emph{Medical Physics},
  vol.~22, no.~10, pp. 1555--1567, 1995.

\bibitem{lo95}
S.-C.~B. Lo, S.-L. Lou, J.-S. Lin, M.~T. Freedman, M.~V. Chien, S.~K. Mun
  \emph{et~al.}, ``Artificial convolution neural network techniques and
  applications for lung nodule detection,'' \emph{Medical Imaging, IEEE
  Transactions on}, vol.~14, no.~4, pp. 711--718, 1995.

\bibitem{tajbakhsh15b}
N.~Tajbakhsh, S.~R. Gurudu, and J.~Liang, ``A comprehensive computer-aided
  polyp detection system for colonoscopy videos,'' in \emph{Information
  Processing in Medical Imaging}.\hskip 1em plus 0.5em minus 0.4em\relax
  Springer, 2015, pp. 327--338.

\bibitem{tajbakhsh15a}
------, ``Automatic polyp detection in colonoscopy videos using an ensemble of
  convolutional neural networks,'' in \emph{Biomedical Imaging (ISBI), 2015
  IEEE 12th International Symposium on}.\hskip 1em plus 0.5em minus 0.4em\relax
  IEEE, 2015, pp. 79--83.

\bibitem{tajbakhsh15c}
N.~Tajbakhsh and J.~Liang, ``Computer-aided pulmonary embolism detection using
  a novel vessel-aligned multi-planar image representation and convolutional
  neural networks,'' in \emph{Medical Image Computing and Computer-Assisted
  Intervention – MICCAI 2015}, 2015.

\bibitem{cirecsan13}
D.~C. Cire{\c{s}}an, A.~Giusti, L.~M. Gambardella, and J.~Schmidhuber,
  ``Mitosis detection in breast cancer histology images with deep neural
  networks,'' in \emph{Medical Image Computing and Computer-Assisted
  Intervention--MICCAI 2013}.\hskip 1em plus 0.5em minus 0.4em\relax Springer,
  2013, pp. 411--418.

\bibitem{roth14}
H.~Roth, L.~Lu, A.~Seff, K.~Cherry, J.~Hoffman, S.~Wang, J.~Liu, E.~Turkbey,
  and R.~Summers, ``\BIBforeignlanguage{English}{A new 2.5d representation for
  lymph node detection using random sets of deep convolutional neural network
  observations},'' in \emph{\BIBforeignlanguage{English}{Medical Image
  Computing and Computer-Assisted Intervention – MICCAI 2014}}, ser. Lecture
  Notes in Computer Science, P.~Golland, N.~Hata, C.~Barillot, J.~Hornegger,
  and R.~Howe, Eds.\hskip 1em plus 0.5em minus 0.4em\relax Springer
  International Publishing, 2014, vol. 8673, pp. 520--527.

\bibitem{zheng15}
Y.~Zheng, D.~Liu, B.~Georgescu, H.~Nguyen, and D.~Comaniciu, ``3d deep learning
  for efficient and robust landmark detection in volumetric data,'' in
  \emph{Medical Image Computing and Computer-Assisted Intervention – MICCAI
  2015}, 2015.

\bibitem{shin16}
J.~Y. Shin, N.~Tajbakhsh, R.~T. Hurst, C.~B. Kendall, and J.~Liang,
  ``Automating carotid intima-media thickness video interpretation with
  convolutional neural networks,'' to appear in Proceedings of the IEEE
  Conference on Computer Vision and Pattern Recognition, 2016.

\bibitem{roth15a}
H.~R. Roth, A.~Farag, L.~Lu, E.~B. Turkbey, and R.~M. Summers, ``Deep
  convolutional networks for pancreas segmentation in ct imaging,'' in
  \emph{SPIE Medical Imaging}.\hskip 1em plus 0.5em minus 0.4em\relax
  International Society for Optics and Photonics, 2015, pp. 94\,131G--94\,131G.

\bibitem{havaei15}
M.~Havaei, A.~Davy, D.~Warde-Farley, A.~Biard, A.~Courville, Y.~Bengio, C.~Pal,
  P.-M. Jodoin, and H.~Larochelle, ``Brain tumor segmentation with deep neural
  networks,'' \emph{arXiv preprint arXiv:1505.03540}, 2015.

\bibitem{zhang15}
W.~Zhang, R.~Li, H.~Deng, L.~Wang, W.~Lin, S.~Ji, and D.~Shen, ``Deep
  convolutional neural networks for multi-modality isointense infant brain
  image segmentation,'' \emph{NeuroImage}, vol. 108, pp. 214--224, 2015.

\bibitem{ciresan12}
D.~Ciresan, A.~Giusti, L.~M. Gambardella, and J.~Schmidhuber, ``Deep neural
  networks segment neuronal membranes in electron microscopy images,'' in
  \emph{Advances in Neural Information Processing Systems 25}, F.~Pereira,
  C.~Burges, L.~Bottou, and K.~Weinberger, Eds.\hskip 1em plus 0.5em minus
  0.4em\relax Curran Associates, Inc., 2012, pp. 2843--2851.

\bibitem{prasoon13}
A.~Prasoon, K.~Petersen, C.~Igel, F.~Lauze, E.~Dam, and M.~Nielsen, ``Deep
  feature learning for knee cartilage segmentation using a triplanar
  convolutional neural network,'' in \emph{Medical Image Computing and
  Computer-Assisted Intervention--MICCAI 2013}.\hskip 1em plus 0.5em minus
  0.4em\relax Springer, 2013, pp. 246--253.

\bibitem{bar15}
Y.~Bar, I.~Diamant, L.~Wolf, and H.~Greenspan, ``Deep learning with non-medical
  training used for chest pathology identification,'' in \emph{SPIE Medical
  Imaging}.\hskip 1em plus 0.5em minus 0.4em\relax International Society for
  Optics and Photonics, 2015, pp. 94\,140V--94\,140V.

\bibitem{Ginneken15}
B.~van Ginneken, A.~A. Setio, C.~Jacobs, and F.~Ciompi, ``Off-the-shelf
  convolutional neural network features for pulmonary nodule detection in
  computed tomography scans,'' in \emph{Biomedical Imaging (ISBI), 2015 IEEE
  12th International Symposium on}, April 2015, pp. 286--289.

\bibitem{Arevalo15}
J.~Arevalo, F.~Gonzalez, R.~Ramos-Pollan, J.~Oliveira, and M.~Guevara~Lopez,
  ``Convolutional neural networks for mammography mass lesion classification,''
  in \emph{Engineering in Medicine and Biology Society (EMBC), 2015 37th Annual
  International Conference of the IEEE}, Aug 2015, pp. 797--800.

\bibitem{schlegl14}
T.~Schlegl, J.~Ofner, and G.~Langs, ``Unsupervised pre-training across image
  domains improves lung tissue classification,'' in \emph{Medical Computer
  Vision: Algorithms for Big Data}.\hskip 1em plus 0.5em minus 0.4em\relax
  Springer, 2014, pp. 82--93.

\bibitem{chen15}
H.~Chen, D.~Ni, J.~Qin, S.~Li, X.~Yang, T.~Wang, and P.~A. Heng, ``Standard
  plane localization in fetal ultrasound via domain transferred deep neural
  networks,'' \emph{Biomedical and Health Informatics, IEEE Journal of},
  vol.~19, no.~5, pp. 1627--1636, Sept 2015.

\bibitem{Gustavo15}
\BIBentryALTinterwordspacing
G.~Carneiro, J.~Nascimento, and A.~Bradley,
  ``\BIBforeignlanguage{English}{Unregistered multiview mammogram analysis with
  pre-trained deep learning models},'' in
  \emph{\BIBforeignlanguage{English}{Medical Image Computing and
  Computer-Assisted Intervention – MICCAI 2015}}, ser. Lecture Notes in
  Computer Science, N.~Navab, J.~Hornegger, W.~M. Wells, and A.~F. Frangi,
  Eds.\hskip 1em plus 0.5em minus 0.4em\relax Springer International
  Publishing, 2015, vol. 9351, pp. 652--660. [Online]. Available:
  \url{http://dx.doi.org/10.1007/978-3-319-24574-4_78}
\BIBentrySTDinterwordspacing

\bibitem{shin15}
H.-C. Shin, L.~Lu, L.~Kim, A.~Seff, J.~Yao, and R.~M. Summers, ``Interleaved
  text/image deep mining on a very large-scale radiology database,'' in
  \emph{Proceedings of the IEEE Conference on Computer Vision and Pattern
  Recognition}, 2015, pp. 1090--1099.

\bibitem{Gao15}
M.~Gao, U.~Bagci, L.~Lu, A.~Wu, M.~Buty, H.-C. Shin, H.~Roth, G.~Z. Papadakis,
  A.~Depeursinge, R.~M. Summers \emph{et~al.}, ``Holistic classification of ct
  attenuation patterns for interstitial lung diseases via deep convolutional
  neural networks,'' in \emph{the 1st Workshop on Deep Learning in Medical
  Image Analysis, International Conference on Medical Image Computing and
  Computer Assisted Intervention, at MICCAI-DLMIA'15}, 2015.

\bibitem{margeta15}
J.~Margeta, A.~Criminisi, R.~Cabrera~Lozoya, D.~C. Lee, and N.~Ayache,
  ``Fine-tuned convolutional neural nets for cardiac mri acquisition plane
  recognition,'' \emph{Computer Methods in Biomechanics and Biomedical
  Engineering: Imaging \& Visualization}, pp. 1--11, 2015.

\bibitem{hubel59}
D.~H. Hubel and T.~N. Wiesel, ``Receptive fields of single neurones in the
  cat's striate cortex,'' \emph{The Journal of physiology}, vol. 148, no.~3,
  pp. 574--591, 1959.

\bibitem{Edwards02}
D.~C. Edwards, M.~A. Kupinski, C.~E. Metz, and R.~M. Nishikawa, ``Maximum
  likelihood fitting of {FROC} curves under an
  initial-detection-and-candidate-analysis model,'' \emph{Medical physics},
  vol.~29, no.~12, pp. 2861--2870, 2002.

\bibitem{jia14}
Y.~Jia, E.~Shelhamer, J.~Donahue, S.~Karayev, J.~Long, R.~Girshick,
  S.~Guadarrama, and T.~Darrell, ``Caffe: Convolutional architecture for fast
  feature embedding,'' \emph{arXiv preprint arXiv:1408.5093}, 2014.

\bibitem{glorot10}
X.~Glorot and Y.~Bengio, ``Understanding the difficulty of training deep
  feedforward neural networks,'' in \emph{International conference on
  artificial intelligence and statistics}, 2010, pp. 249--256.

\bibitem{he15}
K.~He, X.~Zhang, S.~Ren, and J.~Sun, ``Delving deep into rectifiers: Surpassing
  human-level performance on imagenet classification,'' \emph{arXiv preprint
  arXiv:1502.01852}, 2015.

\bibitem{tajbakhsh15d}
N.~Tajbakhsh, S.~Gurudu, and J.~Liang, ``Automated polyp detection in
  colonoscopy videos using shape and context information,'' \emph{Medical
  Imaging, IEEE Transactions on}, vol.~PP, no.~99, pp. 1--1, 2015.

\bibitem{Pabby05}
A.~Pabby, R.~E. Schoen, J.~L. Weissfeld, R.~Burt, J.~W. Kikendall, P.~Lance,
  M.~Shike, E.~Lanza, and A.~Schatzkin, ``Analysis of colorectal cancer
  occurrence during surveillance colonoscopy in the dietary polyp prevention
  trial,'' \emph{Gastrointest Endosc}, vol.~61, no.~3, pp. 385--91, 2005.

\bibitem{Rijn06}
J.~van Rijn, J.~Reitsma, J.~Stoker, P.~Bossuyt, S.~van Deventer, and E.~Dekker,
  ``Polyp miss rate determined by tandem colonoscopy: a systematic review,''
  \emph{American Journal of Gastroenterology}, vol. 101, no.~2, pp. 343--350,
  2006.

\bibitem{Kim07}
D.~H. Kim, P.~J. Pickhardt, A.~J. Taylor, W.~K. Leung, T.~C. Winter, J.~L.
  Hinshaw, D.~V. Gopal, M.~Reichelderfer, R.~H. Hsu, and P.~R. Pfau, ``Ct
  colonography versus colonoscopy for the detection of advanced neoplasia,''
  \emph{N Engl J Med}, vol. 357, no.~14, pp. 1403--12, 2007.

\bibitem{Heresbach08}
D.~Heresbach, T.~Barrioz, M.~Lapalus, D.~Coumaros, P.~Bauret, P.~Potier,
  D.~Sautereau, C.~Bousti{\`e}re, J.~Grimaud, C.~Barth{\'e}l{\'e}my
  \emph{et~al.}, ``Miss rate for colorectal neoplastic polyps: a prospective
  multicenter study of back-to-back video colonoscopies.'' \emph{Endoscopy},
  vol.~40, no.~4, pp. 284--290, 2008.

\bibitem{leufkens12}
A.~Leufkens, M.~van Oijen, F.~Vleggaar, and P.~Siersema, ``Factors influencing
  the miss rate of polyps in a back-to-back colonoscopy study,''
  \emph{Endoscopy}, vol.~44, no.~05, pp. 470--475, 2012.

\bibitem{Rabeneck03}
L.~Rabeneck, H.~El-Serag, J.~Davila, and R.~Sandler, ``Outcomes of colorectal
  cancer in the united states: no change in survival (1986-1997).'' \emph{The
  American journal of gastroenterology}, vol.~98, no.~2, p. 471, 2003.

\bibitem{karkanis03}
S.~A. Karkanis, D.~K. Iakovidis, D.~E. Maroulis, D.~A. Karras, and M.~Tzivras,
  ``Computer-aided tumor detection in endoscopic video using color wavelet
  features,'' \emph{Information Technology in Biomedicine, IEEE Transactions
  on}, vol.~7, no.~3, pp. 141--152, 2003.

\bibitem{iakovidis05}
D.~K. Iakovidis, D.~E. Maroulis, S.~A. Karkanis, and A.~Brokos, ``A comparative
  study of texture features for the discrimination of gastric polyps in
  endoscopic video,'' in \emph{Computer-Based Medical Systems, 2005.
  Proceedings. 18th IEEE Symposium on}.\hskip 1em plus 0.5em minus 0.4em\relax
  IEEE, 2005, pp. 575--580.

\bibitem{alexandre08}
L.~A. Alexandre, N.~Nobre, and J.~Casteleiro, ``Color and position versus
  texture features for endoscopic polyp detection,'' in \emph{BioMedical
  Engineering and Informatics, 2008. BMEI 2008. International Conference on},
  vol.~2.\hskip 1em plus 0.5em minus 0.4em\relax IEEE, 2008, pp. 38--42.

\bibitem{Hwang07}
S.~Hwang, J.~Oh, W.~Tavanapong, J.~Wong, and P.~de~Groen, ``Polyp detection in
  colonoscopy video using elliptical shape feature,'' in \emph{Image
  Processing, 2007. ICIP 2007. IEEE International Conference on}, vol.~2, 2007,
  pp. II--465--II--468.

\bibitem{Bernal12}
J.~Bernal, J.~Sánchez, and F.~Vilariño, ``Towards automatic polyp detection
  with a polyp appearance model,'' \emph{Pattern Recognition}, vol.~45, no.~9,
  pp. 3166--3182, 2012.

\bibitem{Bernal13}
J.~Bernal, J.~S{\'a}nchez, and F.~Vilarino, ``Impact of image preprocessing
  methods on polyp localization in colonoscopy frames,'' in \emph{Engineering
  in Medicine and Biology Society (EMBC), 2013 35th Annual International
  Conference of the IEEE}.\hskip 1em plus 0.5em minus 0.4em\relax IEEE, 2013,
  pp. 7350--7354.

\bibitem{Wang13}
Y.~Wang, W.~Tavanapong, J.~Wong, J.~Oh, and P.~de~Groen, ``Part-based
  multi-derivative edge cross-section profiles for polyp detection in
  colonoscopy,'' \emph{Biomedical and Health Informatics, IEEE Journal of},
  vol.~PP, no.~99, pp. 1--1, 2013.

\bibitem{park12}
S.~Y. Park, D.~Sargent, I.~Spofford, K.~Vosburgh, and Y.~A-Rahim, ``A colon
  video analysis framework for polyp detection,'' \emph{Biomedical Engineering,
  IEEE Transactions on}, vol.~59, no.~5, pp. 1408--1418, 2012.

\bibitem{Nima13}
N.~Tajbakhsh, S.~Gurudu, and J.~Liang, ``A classification-enhanced vote
  accumulation scheme for detecting colonic polyps,'' in \emph{Abdominal
  Imaging. Computation and Clinical Applications}, ser. Lecture Notes in
  Computer Science, 2013, vol. 8198, pp. 53--62.

\bibitem{Nima14p}
N.~Tajbakhsh, C.~Chi, S.~R. Gurudu, and J.~Liang, ``Automatic polyp detection
  from learned boundaries,'' in \emph{Biomedical Imaging (ISBI), 2014 IEEE 10th
  International Symposium on}, 2014.

\bibitem{Nima14p2}
N.~Tajbakhsh, S.~R. Gurudu, and J.~Liang, ``Automatic polyp detection using
  global geometric constraints and local intensity variation patterns,'' in
  \emph{Medical Image Computing and Computer-Assisted Intervention--MICCAI
  2014}.\hskip 1em plus 0.5em minus 0.4em\relax Springer, 2014, pp. 179--187.

\bibitem{liang07}
J.~Liang and J.~Bi, ``Computer aided detection of pulmonary embolism with
  tobogganing and multiple instance classification in {CT} pulmonary
  angiography,'' in \emph{Information Processing in Medical Imaging}.\hskip 1em
  plus 0.5em minus 0.4em\relax Springer, 2007, pp. 630--641.

\bibitem{calderthe2005}
\BIBentryALTinterwordspacing
K.~K. Calder, M.~Herbert, and S.~O. Henderson, ``The mortality of untreated
  pulmonary embolism in emergency department patients.'' \emph{Annals of
  emergency medicine}, vol.~45, no.~3, pp. 302--310, 2005. [Online]. Available:
  \url{http://dx.doi.org/10.1016/j.annemergmed.2004.10.001}
\BIBentrySTDinterwordspacing

\bibitem{sadighchallenges2011}
\BIBentryALTinterwordspacing
G.~Sadigh, A.~M. Kelly, and P.~Cronin, ``Challenges, controversies, and hot
  topics in pulmonary embolism imaging,'' \emph{American Journal of
  Roentgenology}, vol. 196, no.~3, 2011. [Online]. Available:
  \url{http://dx.doi.org/10.2214/AJR.10.5830}
\BIBentrySTDinterwordspacing

\bibitem{fairfield90}
J.~Fairfield, ``Toboggan contrast enhancement for contrast segmentation,'' in
  \emph{Pattern Recognition, 1990. Proceedings., 10th International Conference
  on}, vol.~1.\hskip 1em plus 0.5em minus 0.4em\relax IEEE, 1990, pp. 712--716.

\bibitem{haralick73}
R.~M. Haralick, K.~Shanmugam, and I.~H. Dinstein, ``Textural features for image
  classification,'' \emph{Systems, Man and Cybernetics, IEEE Transactions on},
  no.~6, pp. 610--621, 1973.

\bibitem{tajbakhsh14q}
N.~Tajbakhsh, C.~Chi, H.~Sharma, Q.~Wu, S.~R. Gurudu, and J.~Liang, ``Automatic
  assessment of image informativeness in colonoscopy,'' in \emph{Abdominal
  Imaging. Computational and Clinical Applications}.\hskip 1em plus 0.5em minus
  0.4em\relax Springer, 2014, pp. 151--158.

\bibitem{arnold09}
M.~Arnold, A.~Ghosh, G.~Lacey, S.~Patchett, and H.~Mulcahy, ``Indistinct frame
  detection in colonoscopy videos,'' in \emph{Machine Vision and Image
  Processing Conference, 2009. IMVIP'09. 13th International}.\hskip 1em plus
  0.5em minus 0.4em\relax IEEE, 2009, pp. 47--52.

\bibitem{oh07}
J.~Oh, S.~Hwang, J.~Lee, W.~Tavanapong, J.~Wong, and P.~C. de~Groen,
  ``Informative frame classification for endoscopy video,'' \emph{Medical Image
  Analysis}, vol.~11, no.~2, pp. 110--127, 2007.

\bibitem{menchon15}
R.-M. Mench{\'o}n-Lara and J.-L. Sancho-G{\'o}mez, ``Fully automatic
  segmentation of ultrasound common carotid artery images based on machine
  learning,'' \emph{Neurocomputing}, vol. 151, pp. 161--167, 2015.

\bibitem{menchon13}
R.-M. Mench{\'o}n-Lara, M.-C. Bastida-Jumilla, A.~Gonz{\'a}lez-L{\'o}pez, and
  J.~L. Sancho-G{\'o}mez, ``Automatic evaluation of carotid intima-media
  thickness in ultrasounds using machine learning,'' in \emph{Natural and
  Artificial Computation in Engineering and Medical Applications}.\hskip 1em
  plus 0.5em minus 0.4em\relax Springer, 2013, pp. 241--249.

\bibitem{petroudi12}
S.~Petroudi, C.~Loizou, M.~Pantziaris, and C.~Pattichis, ``Segmentation of the
  common carotid intima-media complex in ultrasound images using active
  contours,'' \emph{Biomedical Engineering, IEEE Transactions on}, vol.~59,
  no.~11, pp. 3060--3069, 2012.

\bibitem{xu12}
X.~Xu, Y.~Zhou, X.~Cheng, E.~Song, and G.~Li, ``Ultrasound intima--media
  segmentation using hough transform and dual snake model,'' \emph{Computerized
  Medical Imaging and Graphics}, vol.~36, no.~3, pp. 248--258, 2012.

\bibitem{liang06}
J.~Liang, T.~McInerney, and D.~Terzopoulos, ``United snakes,'' \emph{Medical
  image analysis}, vol.~10, no.~2, pp. 215--233, 2006.

\bibitem{sharma14}
H.~Sharma, R.~G. Golla, Y.~Zhang, C.~B. Kendall, R.~T. Hurst, N.~Tajbakhsh, and
  J.~Liang, ``Ecg-based frame selection and curvature-based roi detection for
  measuring carotid intima-media thickness,'' in \emph{SPIE Medical
  Imaging}.\hskip 1em plus 0.5em minus 0.4em\relax International Society for
  Optics and Photonics, 2014, pp. 904\,016--904\,016.

\bibitem{lee09}
H.~Lee, R.~Grosse, R.~Ranganath, and A.~Y. Ng, ``Convolutional deep belief
  networks for scalable unsupervised learning of hierarchical
  representations,'' in \emph{Proceedings of the 26th Annual International
  Conference on Machine Learning}.\hskip 1em plus 0.5em minus 0.4em\relax ACM,
  2009, pp. 609--616.

\bibitem{abdel14}
O.~Abdel-Hamid, A.-r. Mohamed, H.~Jiang, L.~Deng, G.~Penn, and D.~Yu,
  ``Convolutional neural networks for speech recognition,'' \emph{Audio,
  Speech, and Language Processing, IEEE/ACM Transactions on}, vol.~22, no.~10,
  pp. 1533--1545, 2014.

\bibitem{wulsin11}
D.~Wulsin, J.~Gupta, R.~Mani, J.~Blanco, and B.~Litt, ``Modeling
  electroencephalography waveforms with semi-supervised deep belief nets: fast
  classification and anomaly measurement,'' \emph{Journal of neural
  engineering}, vol.~8, no.~3, p. 036015, 2011.

\end{thebibliography}

%\hbox{}\newpage
%\hbox{}\newpage

\beginsupplement
\onecolumn
%\vspace{-1cm}
\section*{Supplementary material}
%\vspace{-1cm}

  \begin{table}[h]
\caption{ {\orange Statistical comparisons between the FROC curves shown in \figurename~\ref{fig:Polyp_FROC} for polyp detection (level of  significance is $\alpha=0.05$). The curves are compared at  0.01 and .001 false positives per frame, because they coincide with the elbows of the performance curves where they yield relatively higher sensitivity. A red cell indicates that a pair of curves are statistically different in neither of the  chosen operating point whereas a green cell indicates at which operating points a statistically significant difference is observed. }}
\begin{center}

  \begin{tabular}{*{10}{|c|c|c|c|c|c|c|c|c}}

\hline

%\multirow{ 2}{*}{\multicolumn{10}{c}{PE detection}} \\
%\diagbox[height=2.5cm, width =2cm]{method}{p vale}{method}
& \verText{FT:only fc8} &\verText{FT:fc7-fc8} &\verText{FT:fc6-fc8} & \verText{FT:conv5-fc8}& \verText{FT:conv4-fc8}& \verText{FT:conv3-fc8}& \verText{FT:conv2-fc8}& \verText{FT:conv1-fc8}& \verText{AlexNet scratch}\\
%& & & & & & & & & & &\\
\hline

FT:only fc8& \cellcolor{gray!75}& \cellcolor{gray!75}& \cellcolor{gray!75}& \cellcolor{gray!75}& \cellcolor{gray!75}& \cellcolor{gray!75}& \cellcolor{gray!75}& \cellcolor{gray!75}& \cellcolor{gray!75}\\[.05cm]
FT:fc7-fc8& \cellcolor{green!75}$10^{-2,-3}$& \cellcolor{gray!75}& \cellcolor{gray!75}& \cellcolor{gray!75}& \cellcolor{gray!75}& \cellcolor{gray!75}& \cellcolor{gray!75}& \cellcolor{gray!75}& \cellcolor{gray!75}\\[.05cm]
FT:fc6-fc8& \cellcolor{green!75}$10^{-2,-3}$& \cellcolor{red!75}& \cellcolor{gray!75}& \cellcolor{gray!75}& \cellcolor{gray!75}& \cellcolor{gray!75}& \cellcolor{gray!75}& \cellcolor{gray!75}& \cellcolor{gray!75}\\[.05cm]
FT:conv5-fc8& \cellcolor{green!75}$10^{-2,-3}$& \cellcolor{green!75}$10^{-2,-3}$& \cellcolor{green!75}$10^{-2,-3}$& \cellcolor{gray!75}& \cellcolor{gray!75}& \cellcolor{gray!75}& \cellcolor{gray!75}& \cellcolor{gray!75}& \cellcolor{gray!75}\\[.05cm]
FT:conv4-fc8& \cellcolor{green!75}$10^{-2,-3}$& \cellcolor{green!75}$10^{-2,-3}$& \cellcolor{green!75}$10^{-2,-3}$& \cellcolor{red!75}& \cellcolor{gray!75}& \cellcolor{gray!75}& \cellcolor{gray!75}& \cellcolor{gray!75}& \cellcolor{gray!75}\\[.05cm]
FT:conv3-fc8& \cellcolor{green!75}$10^{-2,-3}$& \cellcolor{green!75}$10^{-2,-3}$& \cellcolor{green!75}$10^{-2,-3}$& \cellcolor{green!75}$10^{-3}$& \cellcolor{red!75}& \cellcolor{gray!75}& \cellcolor{gray!75}& \cellcolor{gray!75}& \cellcolor{gray!75}\\[.05cm]
FT:conv2-fc8& \cellcolor{green!75}$10^{-2,-3}$& \cellcolor{green!75}$10^{-2,-3}$& \cellcolor{green!75}$10^{-2,-3}$& \cellcolor{green!75}$10^{-2,-3}$& \cellcolor{green!75}$10^{-2,-3}$& \cellcolor{green!75}$10^{-2,-3}$& \cellcolor{gray!75}& \cellcolor{gray!75}& \cellcolor{gray!75}\\[.05cm]
FT:conv1-fc8& \cellcolor{green!75}$10^{-2,-3}$& \cellcolor{green!75}$10^{-2,-3}$& \cellcolor{green!75}$10^{-2,-3}$& \cellcolor{green!75}$10^{-2,-3}$& \cellcolor{green!75}$10^{-2,-3}$& \cellcolor{green!75}$10^{-2,-3}$& \cellcolor{red!75}& \cellcolor{gray!75}& \cellcolor{gray!75}\\[.05cm]
AlexNet scratch& \cellcolor{green!75}$10^{-2,-3}$& \cellcolor{green!75}$10^{-2}$& \cellcolor{green!75}$10^{-2,-3}$& \cellcolor{green!75}$10^{-2,-3}$& \cellcolor{green!75}$10^{-2,-3}$& \cellcolor{green!75}$10^{-3}$& \cellcolor{green!75}$10^{-3}$& \cellcolor{green!75}$10^{-2,-3}$& \cellcolor{gray!75}\\[.05cm]
Handcrafted \cite{ tajbakhsh15d}& \cellcolor{green!75}$10^{-2,-3}$& \cellcolor{green!75}$10^{-2,-3}$& \cellcolor{green!75}$10^{-2,-3}$& \cellcolor{green!75}$10^{-2,-3}$& \cellcolor{green!75}$10^{-2,-3}$& \cellcolor{green!75}$10^{-2,-3}$& \cellcolor{green!75}$10^{-2,-3}$& \cellcolor{green!75}$10^{-2,-3}$& \cellcolor{green!75}$10^{-2,-3}$\\[.05cm]

\hline
\end{tabular}
\end{center}
\label{table:polyp_statAnal}
\end{table}

  \begin{table}[h]
\caption{{\orange Statistical comparisons between the FROC curves shown in \figurename~\ref{fig:PE_FROC} for pulmonary embolism detection (level of  significance is $\alpha$=0.05). Each cell presents a statistical comparison between a pair of FROC curves at 1, 2, 3, 4, and 5 false positives per volume. A red cell indicates that the two curves are not statistically different at any of the five operating points, but a green cell contains  the  operating points at which the two curves are statistically different. }}
\begin{center}
  \begin{tabular}{*{10}{|c|c|c|c|c|c|c|c|c}}

\hline

%\multirow{ 2}{*}{\multicolumn{10}{c}{PE detection}} \\
%\diagbox[height=2.5cm, width =2cm]{method}{p vale}{method}
& \verText{FT:only fc8} &\verText{FT:fc7-fc8} &\verText{FT:fc6-fc8} & \verText{FT:conv5-fc8}& \verText{FT:conv4-fc8}& \verText{FT:conv3-fc8}& \verText{FT:conv2-fc8}& \verText{FT:conv1-fc8}& \verText{AlexNet scratch}\\
%& & & & & & & & & & &\\
\hline

FT:only fc8& \cellcolor{gray!75}& \cellcolor{gray!75}& \cellcolor{gray!75}& \cellcolor{gray!75}& \cellcolor{gray!75}& \cellcolor{gray!75}& \cellcolor{gray!75}& \cellcolor{gray!75}& \cellcolor{gray!75}\\[.05cm]
FT:fc7-fc8& \cellcolor{green!75}2,3,4,5& \cellcolor{gray!75}& \cellcolor{gray!75}& \cellcolor{gray!75}& \cellcolor{gray!75}& \cellcolor{gray!75}& \cellcolor{gray!75}& \cellcolor{gray!75}& \cellcolor{gray!75}\\[.05cm]
FT:fc6-fc8& \cellcolor{green!75}1,2,3,4,5& \cellcolor{red!75}& \cellcolor{gray!75}& \cellcolor{gray!75}& \cellcolor{gray!75}& \cellcolor{gray!75}& \cellcolor{gray!75}& \cellcolor{gray!75}& \cellcolor{gray!75}\\[.05cm]
FT:conv5-fc8& \cellcolor{green!75}1,2,3,4,5& \cellcolor{green!75}1,2& \cellcolor{red!75}& \cellcolor{gray!75}& \cellcolor{gray!75}& \cellcolor{gray!75}& \cellcolor{gray!75}& \cellcolor{gray!75}& \cellcolor{gray!75}\\[.05cm]
FT:conv4-fc8& \cellcolor{green!75}1,2,3,4,5& \cellcolor{green!75}1,2,3& \cellcolor{red!75}& \cellcolor{red!75}& \cellcolor{gray!75}& \cellcolor{gray!75}& \cellcolor{gray!75}& \cellcolor{gray!75}& \cellcolor{gray!75}\\[.05cm]
FT:conv3-fc8& \cellcolor{green!75}1,2,3,4,5& \cellcolor{green!75}1,2,3,5& \cellcolor{green!75}1& \cellcolor{red!75}& \cellcolor{red!75}& \cellcolor{gray!75}& \cellcolor{gray!75}& \cellcolor{gray!75}& \cellcolor{gray!75}\\[.05cm]
FT:conv2-fc8& \cellcolor{green!75}1,2,3,4,5& \cellcolor{green!75}1,2,3,4,5& \cellcolor{red!75}& \cellcolor{red!75}& \cellcolor{red!75}& \cellcolor{red!75}& \cellcolor{gray!75}& \cellcolor{gray!75}& \cellcolor{gray!75}\\[.05cm]
FT:conv1-fc8& \cellcolor{green!75}1,2,3,4,5& \cellcolor{green!75}1,2,3,4,5& \cellcolor{green!75}3& \cellcolor{red!75}& \cellcolor{red!75}& \cellcolor{red!75}& \cellcolor{red!75}& \cellcolor{gray!75}& \cellcolor{gray!75}\\[.05cm]
AlexNet scratch& \cellcolor{green!75}1,2,3,4,5& \cellcolor{green!75}1,2,3& \cellcolor{red!75}& \cellcolor{red!75}& \cellcolor{red!75}& \cellcolor{red!75}& \cellcolor{red!75}& \cellcolor{red!75}& \cellcolor{gray!75}\\[.05cm]
Handcrafted \cite{liang07}& \cellcolor{green!75}1,2,3,4,5& \cellcolor{green!75}1,2,3,5& \cellcolor{red!75}& \cellcolor{red!75}& \cellcolor{red!75}& \cellcolor{red!75}& \cellcolor{red!75}& \cellcolor{red!75}& \cellcolor{red!75}\\[.05cm]

\hline
\end{tabular}
\end{center}
\label{table:PE_statAnal}
\end{table}

  \begin{table*}
\caption{{\orange Statistical comparisons between the ROC curves shown in \figurename~\ref{fig:QA_ROC} for frame classification (level of  significance is $\alpha$=0.05). Each cell presents a statistical comparison between a pair of ROC curves at false positive rate of 10\%, 15\%, and 20\% (0.1, 0.15, and 0.2 on the horizontal axis). A red cell indicates that the two curves are not statistically different at any of the two operating points, but a green cell contains  the  operating points at which the two curves are statistically different. }}
\begin{center}
  \begin{tabular}{*{10}{|c|c|c|c|c|c|c|c|c}}

\hline

%\multirow{ 2}{*}{\multicolumn{10}{c}{PE detection}} \\
%\diagbox[height=2.5cm, width =2cm]{method}{p vale}{method}
& \verText{FT:only fc8} &\verText{FT:fc7-fc8} &\verText{FT:fc6-fc8} & \verText{FT:conv5-fc8}& \verText{FT:conv4-fc8}& \verText{FT:conv3-fc8}& \verText{FT:conv2-fc8}& \verText{FT:conv1-fc8}& \verText{AlexNet scratch}\\
%& & & & & & & & & & &\\
\hline

FT:only fc8& \cellcolor{gray!75}& \cellcolor{gray!75}& \cellcolor{gray!75}& \cellcolor{gray!75}& \cellcolor{gray!75}& \cellcolor{gray!75}& \cellcolor{gray!75}& \cellcolor{gray!75}& \cellcolor{gray!75}\\[.05cm]
FT:fc7-fc8& \cellcolor{green!75}0.1& \cellcolor{gray!75}& \cellcolor{gray!75}& \cellcolor{gray!75}& \cellcolor{gray!75}& \cellcolor{gray!75}& \cellcolor{gray!75}& \cellcolor{gray!75}& \cellcolor{gray!75}\\[.05cm]
FT:fc6-fc8& \cellcolor{green!75}0.1& \cellcolor{red!75}& \cellcolor{gray!75}& \cellcolor{gray!75}& \cellcolor{gray!75}& \cellcolor{gray!75}& \cellcolor{gray!75}& \cellcolor{gray!75}& \cellcolor{gray!75}\\[.05cm]
FT:conv5-fc8& \cellcolor{green!75}0.1,0.15& \cellcolor{red!75}& \cellcolor{red!75}& \cellcolor{gray!75}& \cellcolor{gray!75}& \cellcolor{gray!75}& \cellcolor{gray!75}& \cellcolor{gray!75}& \cellcolor{gray!75}\\[.05cm]
FT:conv4-fc8& \cellcolor{green!75}0.1,0.15& \cellcolor{red!75}& \cellcolor{red!75}& \cellcolor{red!75}& \cellcolor{gray!75}& \cellcolor{gray!75}& \cellcolor{gray!75}& \cellcolor{gray!75}& \cellcolor{gray!75}\\[.05cm]
FT:conv3-fc8& \cellcolor{green!75}0.1,0.15& \cellcolor{red!75}& \cellcolor{red!75}& \cellcolor{red!75}& \cellcolor{red!75}& \cellcolor{gray!75}& \cellcolor{gray!75}& \cellcolor{gray!75}& \cellcolor{gray!75}\\[.05cm]
FT:conv2-fc8& \cellcolor{green!75}0.1,0.15& \cellcolor{red!75}& \cellcolor{red!75}& \cellcolor{red!75}& \cellcolor{red!75}& \cellcolor{red!75}& \cellcolor{gray!75}& \cellcolor{gray!75}& \cellcolor{gray!75}\\[.05cm]
FT:conv1-fc8& \cellcolor{green!75}0.1,0.15& \cellcolor{red!75}& \cellcolor{red!75}& \cellcolor{red!75}& \cellcolor{red!75}& \cellcolor{red!75}& \cellcolor{red!75}& \cellcolor{gray!75}& \cellcolor{gray!75}\\[.05cm]
AlexNet scratch& \cellcolor{red!75}& \cellcolor{green!75}0.1& \cellcolor{red!75}& \cellcolor{green!75}0.1,0.15& \cellcolor{green!75}0.1,0.15& \cellcolor{red!75}& \cellcolor{green!75}0.15& \cellcolor{red!75}& \cellcolor{gray!75}\\[.05cm]
Handcrafted~\cite{tajbakhsh14q}& \cellcolor{green!75}0.2& \cellcolor{green!75}0.1,0.15,0.2& \cellcolor{green!75}0.1,0.15,0.2& \cellcolor{green!75}0.1,0.15,0.2& \cellcolor{green!75}0.1,0.15,0.2& \cellcolor{green!75}0.1,0.15,0.2& \cellcolor{green!75}0.1,0.15,0.2& \cellcolor{green!75}0.1,0.15,0.2& \cellcolor{green!75}0.15,0.2\\[.05cm]

\hline
\end{tabular}
\end{center}
\label{table:QA_statAnal}
\end{table*}

  \begin{table*}
\caption{{\orange Statistical comparisons between the boxplots shown in \figurename~\ref{fig:CIMT_boxplot}. The p-values larger than 0.05 are highlighted in red.}}

\begin{center}
  \begin{tabular}{*{10}{|c|c|c|c|c|c|c|c|c}}

\hline
 \multicolumn{10}{|c|}{\large Lumen-intima interface}\\
\hline
%\multirow{ 2}{*}{\multicolumn{10}{c}{PE detection}} \\
%\diagbox[height=2.5cm, width =2cm]{method}{p vale}{method}
& \verText{FT:only fc8} &\verText{FT:fc7-fc8} &\verText{FT:fc6-fc8} & \verText{FT:conv5-fc8}& \verText{FT:conv4-fc8}& \verText{FT:conv3-fc8}& \verText{FT:conv2-fc8}& \verText{FT:conv1-fc8}& \verText{AlexNet scratch}\\
%& & & & & & & & & & &\\
\hline

FT:only fc8& \cellcolor{gray!75}& \cellcolor{gray!75}& \cellcolor{gray!75}& \cellcolor{gray!75}& \cellcolor{gray!75}& \cellcolor{gray!75}& \cellcolor{gray!75}& \cellcolor{gray!75}& \cellcolor{gray!75}\\[.05cm]
FT:fc7-fc8&\cellcolor{green!75} p$<$.0001& \cellcolor{gray!75}& \cellcolor{gray!75}& \cellcolor{gray!75}& \cellcolor{gray!75}& \cellcolor{gray!75}& \cellcolor{gray!75}& \cellcolor{gray!75}& \cellcolor{gray!75}\\[.05cm]
FT:fc6-fc8&\cellcolor{green!75} p$<$.0001&\cellcolor{green!75} p$<$.0001& \cellcolor{gray!75}& \cellcolor{gray!75}& \cellcolor{gray!75}& \cellcolor{gray!75}& \cellcolor{gray!75}& \cellcolor{gray!75}& \cellcolor{gray!75}\\[.05cm]
FT:conv5-fc8&\cellcolor{green!75} p$<$.0001&\cellcolor{green!75} p$<$.0001&\cellcolor{green!75} p$<$.001& \cellcolor{gray!75}& \cellcolor{gray!75}& \cellcolor{gray!75}& \cellcolor{gray!75}& \cellcolor{gray!75}& \cellcolor{gray!75}\\[.05cm]
FT:conv4-fc8&\cellcolor{green!75} p$<$.0001&\cellcolor{green!75} p$<$.0001&\cellcolor{green!75} p$<$.0001&\cellcolor{red!75} 0.5808& \cellcolor{gray!75}& \cellcolor{gray!75}& \cellcolor{gray!75}& \cellcolor{gray!75}& \cellcolor{gray!75}\\[.05cm]
FT:conv3-fc8&\cellcolor{green!75} p$<$.0001&\cellcolor{green!75} p$<$.0001&\cellcolor{green!75} p$<$.0001&\cellcolor{red!75} 0.0638&\cellcolor{red!75} 0.0758& \cellcolor{gray!75}& \cellcolor{gray!75}& \cellcolor{gray!75}& \cellcolor{gray!75}\\[.05cm]
FT:conv2-fc8&\cellcolor{green!75} p$<$.0001&\cellcolor{green!75} p$<$.0001&\cellcolor{green!75} p$<$.0001&\cellcolor{red!75} 0.2501&\cellcolor{red!75} 0.3570&\cellcolor{red!75} 0.2284& \cellcolor{gray!75}& \cellcolor{gray!75}& \cellcolor{gray!75}\\[.05cm]
FT:conv1-fc8&\cellcolor{green!75} p$<$.0001&\cellcolor{green!75} p$<$.0001&\cellcolor{green!75} p$<$.0001&\cellcolor{red!75} 0.4183&\cellcolor{red!75} 0.5491&\cellcolor{red!75} 0.4650&\cellcolor{red!75} 0.9530& \cellcolor{gray!75}& \cellcolor{gray!75}\\[.05cm]
AlexNet scratch&\cellcolor{green!75} p$<$.0001&\cellcolor{green!75} p$<$.0001&\cellcolor{red!75} 0.7829&\cellcolor{green!75} p$<$.05&\cellcolor{green!75} p$<$.0001&\cellcolor{green!75} p$<$.0001&\cellcolor{green!75} p$<$.0001&\cellcolor{green!75} p$<$.0001& \cellcolor{gray!75}\\[.05cm]
handCrafted&\cellcolor{red!75} 0.8148&\cellcolor{green!75} p$<$.0001&\cellcolor{green!75} p$<$.0001&\cellcolor{green!75} p$<$.0001&\cellcolor{green!75} p$<$.0001&\cellcolor{green!75} p$<$.0001&\cellcolor{green!75} p$<$.0001&\cellcolor{green!75} p$<$.0001&\cellcolor{green!75} p$<$.0001\\[.05cm]

\end{tabular}

  \begin{tabular}{*{10}{|c|c|c|c|c|c|c|c|c}}

\hline
 \multicolumn{10}{|c|}{\large  Media-adventitia interface}\\
\hline
%\multirow{ 2}{*}{\multicolumn{10}{c}{PE detection}} \\
%\diagbox[height=2.5cm, width =2cm]{method}{p vale}{method}
& \verText{FT:only fc8} &\verText{FT:fc7-fc8} &\verText{FT:fc6-fc8} & \verText{FT:conv5-fc8}& \verText{FT:conv4-fc8}& \verText{FT:conv3-fc8}& \verText{FT:conv2-fc8}& \verText{FT:conv1-fc8}& \verText{AlexNet scratch}\\
%& & & & & & & & & & &\\
\hline

FT:only fc8& \cellcolor{gray!75}& \cellcolor{gray!75}& \cellcolor{gray!75}& \cellcolor{gray!75}& \cellcolor{gray!75}& \cellcolor{gray!75}& \cellcolor{gray!75}& \cellcolor{gray!75}& \cellcolor{gray!75}\\[.05cm]
FT:fc7-fc8& \cellcolor{green!75} p$<$.0001& \cellcolor{gray!75}& \cellcolor{gray!75}& \cellcolor{gray!75}& \cellcolor{gray!75}& \cellcolor{gray!75}& \cellcolor{gray!75}& \cellcolor{gray!75}& \cellcolor{gray!75}\\[.05cm]
FT:fc6-fc8& \cellcolor{green!75} p$<$.0001& \cellcolor{green!75} p$<$.0001& \cellcolor{gray!75}& \cellcolor{gray!75}& \cellcolor{gray!75}& \cellcolor{gray!75}& \cellcolor{gray!75}& \cellcolor{gray!75}& \cellcolor{gray!75}\\[.05cm]
FT:conv5-fc8& \cellcolor{green!75} p$<$.0001& \cellcolor{green!75} p$<$.0001& \cellcolor{green!75} p$<$.05& \cellcolor{gray!75}& \cellcolor{gray!75}& \cellcolor{gray!75}& \cellcolor{gray!75}& \cellcolor{gray!75}& \cellcolor{gray!75}\\[.05cm]
FT:conv4-fc8& \cellcolor{green!75} p$<$.0001& \cellcolor{green!75} p$<$.0001& \cellcolor{green!75} p$<$.0001& \cellcolor{green!75} p$<$.05& \cellcolor{gray!75}& \cellcolor{gray!75}& \cellcolor{gray!75}& \cellcolor{gray!75}& \cellcolor{gray!75}\\[.05cm]
FT:conv3-fc8& \cellcolor{green!75} p$<$.0001& \cellcolor{green!75} p$<$.0001& \cellcolor{green!75} p$<$.05& \cellcolor{red!75}0.7904& \cellcolor{green!75} p$<$.05& \cellcolor{gray!75}& \cellcolor{gray!75}& \cellcolor{gray!75}& \cellcolor{gray!75}\\[.05cm]
FT:conv2-fc8& \cellcolor{green!75} p$<$.0001& \cellcolor{green!75} p$<$.0001& \cellcolor{green!75} p$<$.05& \cellcolor{red!75}0.7160& \cellcolor{green!75} p$<$.05& \cellcolor{red!75}0.8854& \cellcolor{gray!75}& \cellcolor{gray!75}& \cellcolor{gray!75}\\[.05cm]
FT:conv1-fc8& \cellcolor{green!75} p$<$.0001& \cellcolor{green!75} p$<$.0001& \cellcolor{green!75} p$<$.001& \cellcolor{red!75}0.2474& \cellcolor{red!75}0.2456& \cellcolor{red!75}0.2915& \cellcolor{red!75}0.2313& \cellcolor{gray!75}& \cellcolor{gray!75}\\[.05cm]
AlexNet scratch& \cellcolor{green!75} p$<$.0001& \cellcolor{green!75} p$<$.0001& \cellcolor{red!75}0.3954& \cellcolor{red!75}0.2106& \cellcolor{green!75} p$<$.05& \cellcolor{red!75}0.1369& \cellcolor{red!75}0.0981& \cellcolor{green!75} p$<$.05& \cellcolor{gray!75}\\[.05cm]
handCrafted& \cellcolor{red!75}0.5109& \cellcolor{green!75} p$<$.0001& \cellcolor{green!75} p$<$.0001& \cellcolor{green!75} p$<$.0001& \cellcolor{green!75} p$<$.0001& \cellcolor{green!75} p$<$.0001& \cellcolor{green!75} p$<$.0001& \cellcolor{green!75} p$<$.0001& \cellcolor{green!75} p$<$.0001\\[.05cm]

\hline
\end{tabular}
\end{center}
\label{table:interface_pvalues}
\end{table*}
\end{document}